\documentclass[Afour,sageh,times]{sagej}

\usepackage{moreverb,url}

\usepackage[colorlinks,bookmarksopen,bookmarksnumbered,citecolor=red,urlcolor=red]{hyperref}

\newcommand\BibTeX{{\rmfamily B\kern-.05em \textsc{i\kern-.025em b}\kern-.08em
T\kern-.1667em\lower.7ex\hbox{E}\kern-.125emX}}

\def\volumeyear{2016}

\usepackage{amsmath}
\usepackage{amssymb} 
\usepackage{amsfonts} 
\usepackage{mathtools}
\usepackage{bm}
\usepackage{amsthm}

\newtheorem{theorem}{Theorem}

\newtheorem{remark}{\textbf{Remark}}

\usepackage[linesnumbered,ruled,vlined]{algorithm2e}

\SetCommentSty{mycommfont}
\SetKwInput{KwInput}{Input}
\SetKwInput{KwOutput}{Output}

\usepackage{graphicx}
\usepackage{caption}
\usepackage{subcaption}  
\usepackage{colortbl}
\usepackage[most]{tcolorbox}

\usepackage{siunitx}
\usepackage{textcomp}

\usepackage{empheq}

\begin{document}
\setcounter{secnumdepth}{3}

\runninghead{Jordana et al.}

\title{An Introduction to Zero-Order Optimization Techniques for Robotics}

\author{Armand Jordana\affilnum{1}, Jianghan Zhang\affilnum{1}, Joseph Amigo\affilnum{1} and Ludovic Righetti\affilnum{1,}\affilnum{2}}

\affiliation{\affilnum{1}Machines in Motion Laboratory, New York University, Brooklyn, USA.\\
\affilnum{2}Artificial and Natural Intelligence Toulouse Institute (ANITI), Toulouse, France.}

\corrauth{Armand Jordana,
Machines in Motion Laboratory,
NYU Tandon School of Engineering, 
6 MetroTech Center, Brooklyn, NY 11201, USA}

\email{armand.jordana@nyu.edu}

\begin{abstract}
Zero-order optimization techniques are becoming increasingly popular in robotics due to their ability to handle non-differentiable functions and escape local minima. 
These advantages make them particularly useful for trajectory optimization and policy optimization. 
In this work, we propose a mathematical tutorial on random search. It offers a simple and unifying perspective for understanding a wide range of algorithms commonly used in robotics. Leveraging this viewpoint, we classify many trajectory optimization methods under a common framework and derive novel competitive RL algorithms. 
\end{abstract}


\maketitle

\section{Introduction}

In recent years, zero-order (or derivative-free) optimization techniques~\citep{conn2009introduction} have gained a lot of popularity in the robotics community. While zero-order optimization is a well-established field, its widespread deployment in robotics has only been made possible by recent advances in parallel computing and GPU hardware. These improvements have made it possible to deploy sampling-based Model Predictive Control~(MPC) on complex robotic systems~\citep{li2024drop, xue2024full}. In parallel, Reinforcement Learning (RL) has emerged as a powerful tool and has demonstrated state-of-the-art capabilities in locomotion or manipulation~\citep{handa2023dextreme, hoeller2024anymal}. 

While these approaches may appear unrelated at first, they share a common property: they do not require access to the simulator's gradients. 
This characteristic allows them to optimize non-smooth objective functions, which typically arise in problems involving contact, such as locomotion or manipulation. Another benefit is that it avoids the tedious development of efficient gradient implementations. Furthermore, while deterministic gradient-based algorithms, such as gradient descent or Newton's method, are prone to getting stuck in local minima, zero-order techniques are mostly stochastic, which can help escape local minima.

Zero-order optimization has a long history~\citep{matyas1965random}. In its most generic form, zero-order optimization performs a random search by generating samples at random and updating an estimate of the optimal solution based on the performance of each sample.
In the 1970s, Evolutionary Strategies~(ES) gained popularity~\citep{schwefel1993evolution} and later culminated with the CMA-ES algorithm~\citep{hansen2001completely}. 
In trajectory optimization (TO), predictive sampling~\citep{howell2022predictive} is a well-known and simple random search technique. Another popular algorithm is MPPI~\citep{williams2016aggressive}, which was initially derived using information-theoretic arguments.
In RL, Reinforce~\citep{suttonEdition2} was one of the first major algorithms proposed for continuous action space problems.
These algorithms have since inspired many variations.

While most of these algorithms have strong theoretical foundations, this is not always widely recognized within the robotics community. One reason for this is that these algorithms often lie at the intersection of many fields, such as optimization, statistics, machine learning, and control. 
By bringing together results from these areas, our goal is both to bring to light the theoretical properties of existing algorithms and to leverage this understanding to design novel algorithms for robotics.

Recently, \cite{nesterov2017random} introduced Gaussian smoothing and understood random search from an optimization perspective. Inspired by this work, we propose a unified perspective to understand the zero-order optimization techniques that are popular in the robotics community. More specifically, we are interested in algorithms that can find solutions to the following problem:
\begin{align}
    \min_{x \in \mathbb R^n} f(x) .  \label{eq:main_problem}
\end{align}
This formulation encompasses a wide range of problems encountered in robotics, such as TO and RL.
Both of these techniques define a high-level goal through a cost function that has to be minimized (or a reward function that has to be maximized). However, they differ in the optimization space. TO optimizes directly over trajectories (i.e., a finite-dimensional sequence of vectors), whereas RL optimizes over policy parameters (e.g., the weights and biases of a neural network).
Furthermore, they differ in the way they are deployed.
These algorithms can be used either online (i.e., executed at runtime) or offline. On one hand, TO can either be solved online in an MPC loop or solved offline to generate data to train a policy. On the other hand, recent RL successes relied heavily on offline training in simulation~\citep{handa2023dextreme, hoeller2024anymal}.
While these differences may seem fundamental, we show that the algorithms used to solve these problems share many conceptual similarities.

In the trajectory optimization community, several works have studied derivative-free algorithms and established connections among them. For instance, \cite{stulp2012path, bhardwaj2022storm} show how it is possible to combine MPPI and the well-known Covariance Matrix Adaptation (CMA) scheme. Then, \cite{wagener2019online} understood MPPI as a mirror descent algorithm using the natural gradient, and \cite{xue2024full} connected MPPI to diffusion models. Our work builds on these insights and proposes a unifying framework to understand these algorithms. In particular, we formally justify the block-diagonal implementation of the CMA update in the context of TO. In addition, we shed light on the connection between MPPI and recent results showcasing the benefits of the log-sum-exp transform~\citep{scaman2020simple}. Lastly, we benchmark several state-of-the-art approaches on various robotic examples.

In the RL community, \citet{salimans2017evolution, mania2018simple} first explored the use of random search in the parameter space of policies as an alternative to traditional RL algorithms. In this work, we instead investigate how random search techniques can explain the success of popular RL algorithms. More recently, \citet{suh2022bundled, le2024leveraging} drew the connection between policy gradient techniques and Nesterov's random search~\citep{nesterov2017random}. However, these works used this insight to derive TO algorithms based on random search. In contrast, we study how to leverage this understanding to derive novel RL algorithms. Lastly, we refer the reader to \cite{sigaud2019policy, sigaud2023combining} for comprehensive surveys on the connection between RL and ES techniques. Overall, a broad perspective connecting gradient-free approaches in TO and RL is missing. 

To bridge this gap, we propose a mathematical introduction to zero-order optimization algorithms used in robotics. The main contributions of this work are as follows:
\begin{itemize}
    \item A unified treatment of sampling-based methods from TO and RL: we provide a simple way to understand methods such as MPPI, Covariance Matrix Adaptation (CMA), and RL policy gradient methods through the lens of approximate gradient descent.
    \item A theorem connecting the block-diagonal version of CMA to the natural gradient in the context of TO.
    \item Novel RL algorithms: we show how this unified view allows us to naturally derive novel competitive methods as a byproduct.
    \item Empirical comparisons: we benchmark several key methods in the contexts of TO and RL.
\end{itemize}

Section~\ref{section:random_search} introduces the main concepts of random search and sampling-based gradient approximation. We  also discuss the theoretical concepts explaining why these stochastic algorithms are well-suited to avoid getting stuck in local minima. Section~\ref{section:TO} connects popular TO algorithms to random search.
Section~\ref{section:RL} investigates the connections between Gaussian smoothing and RL. Section~\ref{section:population} discusses how to improve the sample efficiency of these algorithms by relying on a population of samples.
Lastly, Section~\ref{section:parralel} discusses practical considerations.

\section{Random search}\label{section:random_search}

In this section, we introduce how a population of samples (or particles) can be used to search for global solutions. The algorithms presented here will serve as building blocks for the understanding of widely used TO and RL algorithms. Depending on the context, the samples may correspond to trajectories in TO or to policy parameters in RL.

\subsection{Simple random search}

First, we study two well-known random search approaches: global search and local search algorithms. Although rarely used in practice, their convergence guarantees help to understand more sophisticated algorithms. 

\subsubsection{Pure (or Global) Random Search}

The most naive random search approach consists of iteratively sampling points at random. If the sampled point has a value lower than the current best, it is kept as the new best guess; otherwise, it is discarded. 
This method, known as Pure Random Search, is summarized in Algorithm~\ref{algo:blind_search}.

\begin{algorithm}[!ht]
\DontPrintSemicolon
\KwInput{$x_0 \in \mathbb R^n$}
$ x \gets x_0$\;
\While{stopping criterion is not met}{
Sample $\tilde{x}$ in $\mathbb R^n$\;
  \If{$f(\tilde{x}) < f(x)$ }{
    $x \gets \tilde{x} $\;
  }
}
\KwOutput{$x$}
\caption{Pure Random Search}\label{algo:blind_search}
\end{algorithm}

Interestingly, the sampling distribution of Algorithm~\ref{algo:blind_search} ignores all the previous estimates. For this reason, it is sometimes referred to as a global algorithm. 
Under some mild conditions on the objective function, Algorithm~\ref{algo:blind_search} converges to a global minimum~\citep{spall2005introduction}. 
Unfortunately, in practice, this method performs poorly as it is subject to the curse of dimensionality. That is to say, the number of function evaluations required to find a global solution grows exponentially with the dimension of the problem.

\subsubsection{Greedy Local Search}

A common attempt to break the curse of dimensionality is to exploit problem-specific knowledge. For instance, exploiting previous information about the shape of the function can help design an empirically more efficient algorithm. This is called Greedy Local Search. The idea is to iteratively search around the previous estimate with additive Gaussian noise, for instance.
Then, as in the previous algorithm, the sample is kept only if it improves the function value. Algorithm~\ref{algo:local_search} summarizes the procedure.

\begin{algorithm}[!ht]
\DontPrintSemicolon
\KwInput{$x_0 \in \mathbb R^n$, $\Sigma$}
$ x \gets x_0$\;
\While{stopping criterion is not met}{
Sample $d$ with $\mathcal{N}(0, \Sigma)$ \; 
  \If{$ f(x + d) < f(x)$ }{
    $x \gets x + d $\;
  }
}
\KwOutput{$x$}
\caption{Greedy Local Search}\label{algo:local_search}
\end{algorithm}

By design, Algorithm~\ref{algo:local_search} depends on previous estimates. For that reason, it can be referred to as a local algorithm. However, this notion should not be confused with the concepts of local and global solutions. In fact, although the convergence guarantees of Algorithm~\ref{algo:local_search} are more restrictive than those of Algorithm~\ref{algo:blind_search}, it is still possible to prove the convergence of Algorithm~\ref{algo:local_search} to a global solution under some regularity conditions~\citep{spall2005introduction}. 
Consequently, both local and global searches can have global convergence guarantees. However, while both are subject to the curse of dimensionality, local search techniques can be more effective empirically. In fact, as we show next, it is well known that local search techniques relying on gradient information can exhibit strong convergence rates and guarantees in convex settings.

\subsection{Random search via gradient approximation}

In this section, we present various ways to perform approximate gradient descent using only function evaluations. In later sections, we show how widely used algorithms in TO and RL can be derived from the basic principles introduced here.
The class of methods under study is summarized in Algorithm~\ref{algo:gradient_descent}. 

\begin{algorithm}[!ht]
\DontPrintSemicolon
\KwInput{$x_0 \in \mathbb R^n$}
$ x \gets x_0$\;
\While{stopping criterion is not met}{
Sample $g$ approximating the gradient at $x$. \tcp{e.g. Eq. \eqref{eq:randomCD}, Eq. \eqref{eq:SPSA}, Eq \eqref{eq:randomized_smoothing} or Eq~\eqref{eq:randomized_smoothing_central} }
$x \gets x - \alpha g $ 
}
\KwOutput{$x$}
\caption{Approximate Gradient Descent}\label{algo:gradient_descent}
\end{algorithm}

Here, $\alpha$ is the step size (or learning rate). There are many ways to estimate the gradient with multiple function evaluations~\citep{berahas2022theoretical}. Among the most widely used techniques is the forward finite difference technique:
\begin{align}
 \sum_{j=1}^n\frac{f(x + \mu e_j) - f(x)}{\mu} e_j, \label{eq:finite_difference}
\end{align}
where $e_j \in \mathbb R^n$ represents the $j^{th}$ canonical vector and where $\mu > 0$ is a scalar. Alternatively, one may use central finite differences. While this estimate can be very effective, it requires $n+1$ function evaluations, which is a limitation in high-dimensional settings. 

In the remainder of this section, we will present multiple ways to approximate the gradient. In other words, we will provide different valid expressions for $g$ in Algorithm \ref{algo:gradient_descent}.

\subsubsection{Random Coordinate Descent}
One way to avoid performing too many function evaluations is Random Coordinate Descent~\citep{wright2015coordinate}. As its name suggests, this method alternates descent steps on random coordinates. 
\begin{empheq}[box=\fbox]{equation}
 g = \frac{f(x + \mu e_j) - f(x)}{\mu} e_j, \label{eq:randomCD}
\end{empheq}
where $j$ can be chosen uniformly at random in $\{1, \dots, n\}$. Note that $g$ depends on $x$.
In the convex setting and without finite-difference approximation (i.e. with the exact gradient), Random Coordinate Descent requires, on average, at most $n$ times more steps compared to classical gradient descent~\citep{wright2015coordinate}. In addition, for a certain class of functions, it can require as many steps as gradient descent while requiring $n$ times fewer function evaluations. Thus, this stochastic estimate offers the hope of matching the gradient's descent performance with a few samples.
Random Coordinate Descent samples discrete coordinates, but it is also possible to sample along arbitrary directions.

\subsubsection{Simultaneous Perturbation Stochastic Approximation (SPSA)}

In the 1990s, Spall introduced Simultaneous Perturbation Stochastic Approximation (SPSA)~\citep{spall2005introduction}. The main idea is to replace finite differences with the following stochastic estimate:
\begin{empheq}[box=\fbox]{equation}
g = \frac{f(x + \mu \Delta) - f(x - \mu \Delta)}{2 \mu }\Delta, \label{eq:SPSA}
\end{empheq}
where each component of \( \Delta \) is equal to $\pm 1$ according to independent Bernoulli distributions. Note that Spall derived SPSA using central differentiation instead of the forward finite difference.
Under reasonable assumptions, SPSA can achieve performance similar to that of finite differences while requiring $n$ times fewer samples~\citep{spall2005introduction}. Hence, similarly to Random Coordinate Descent, this stochastic estimate offers the hope of matching the gradient's descent performance with a few samples.

While both approaches lead to effective algorithms, the next technique provides a more insightful interpretation.

\subsubsection{Gaussian Smoothing}

\cite{nesterov2017random} propose another perspective. Instead of directly approximating the gradients of the original function, they propose studying a smoothed surrogate function
\begin{align}
  f_{\mu}(x) =  \mathbb E [ f(x + \mu \epsilon )],    \label{eq:surrogate-rs}
\end{align}
where $\epsilon\sim \mathcal{N}(0, \Sigma)$. This is Gaussian smoothing, commonly referred to as randomized smoothing (RS). As $\mu$ approaches zero, $f_{\mu}$ tends to $f$. As a result, for small values of $\mu$, this surrogate only slightly changes the objective function (and therefore the minimum location) but adds smoothness~\citep{nesterov2017random}. 
Importantly, the advantage of this surrogate is that its gradient can be evaluated via an average of function evaluations. To understand this, let's first define the constant: $\kappa = \sqrt{(2 \pi)^n \det(\Sigma)}$, the normalization factor of the multivariate Gaussian distribution. With a change of variables (${z = x + \mu \epsilon}$), we can write:
\begin{align}
  f_{\mu}(x) &= \frac{1}{\kappa} \int_{\epsilon} f(x + \mu \epsilon ) e^{-\frac{1}{2}\epsilon^\top \Sigma^{-1}\epsilon} \, d\epsilon \nonumber \\
&= \frac{1}{\kappa \mu^n} \int_z f(z) e^{-\frac{1}{2\mu^2}(z - x)^\top \Sigma^{-1}(z-x)} \, dz.
\end{align}
This allows differentiating $f_{\mu}$ without needing the gradient of the original function,~$f$:
\begin{align}
    \nabla f_{\mu}(x) &= \\
    \frac{1}{\kappa \mu^{n+2}} & \int_z f(z) e^{-\frac{1}{2\mu^2}(z - x)^\top \Sigma^{-1}(z-x)}\Sigma^{-1}(z-x)  dz. \nonumber
\end{align}
With another change of variable ($\epsilon = \mu^{-1} (z-x)$), we have:
\begin{align} 
    \nabla f_{\mu}(x) &= \frac{1}{\kappa \mu} \int_{\epsilon} f(x + \mu \epsilon) e^{-\frac{1}{2}\epsilon^\top \Sigma^{-1}\epsilon}\Sigma^{-1} \epsilon \, d\epsilon \nonumber \\
    &= \mathbb E \left[ \frac{1}{\mu}  f(x + \mu \epsilon ) \Sigma^{-1} \epsilon \right] .
\end{align}

Hence, the gradient of $f_{\mu}$ can be estimated using only function evaluations of the original function, $f$. However, this comes at the cost of approximating an expectation with samples.
Note that the term $\mu^{-1} \Sigma^{-1}\epsilon$ is the gradient of the logarithm of the probability density function with respect to the mean; this is often called the log-likelihood trick. This is sometimes written:
\begin{align}
  \nabla f_{\mu}(x) =  \mathbb E_{z\sim \mathcal{N}(x, \Sigma)} \left[ f(z) \nabla_{m} \log (p_{m = x}(z))\right], \label{eq:loglikelihood_trick0}
\end{align}
where $p_{m}$ is the density of the Gaussian distribution with mean $m$ and covariance $\Sigma$. This alternative view is useful for understanding the CMA algorithm, explained in Section~\ref{sec:CMA}. 
Lastly, since $\mathbb E \left[  f(x) \Sigma^{-1} \epsilon \right] = 0$, we can write:
\begin{align} 
    \nabla f_{\mu}(x) 
    &= \mathbb E \left[ \frac{ f(x + \mu \epsilon )  -  f(x) }{\mu} \Sigma^{-1} \epsilon \right]. \label{eq:randomized_smoothing_Derivation}
\end{align}
In the general case, the expectation is intractable. The idea is then to approximate the gradient of the original function with a stochastic estimate of the gradients of the surrogate. For instance, one can use the following estimate in Algorithm~\ref{algo:gradient_descent}:
\begin{empheq}[box=\fbox]{equation}
  g =  \frac{ f(x + \mu \epsilon )  -  f(x) }{\mu} \Sigma^{-1} \epsilon. \label{eq:randomized_smoothing}
\end{empheq}
While $ \frac{1}{\mu}  f(x + \mu \epsilon ) \Sigma^{-1} \epsilon$ is a valid gradient estimate, it can be arbitrarily large. In contrast, the estimate~\eqref{eq:randomized_smoothing} is invariant to constant translations of the function. Intuitively, this reduces the variance of the estimate.
Note that it is also possible to use an estimate with central differentiation:

\begin{empheq}[box=\fbox]{equation}
  g =  \frac{ f(x + \mu \epsilon )  -  f(x - \mu \epsilon) }{2 \mu} \Sigma^{-1} \epsilon. \label{eq:randomized_smoothing_central}
\end{empheq}

\begin{remark}
Alternatively, one may sample, $\epsilon$, from the surface of a unit sphere~\citep{flaxman2004online}. This technique guarantees that the norm of the direction vectors is bounded.
\end{remark}

In the convex setting, using the forward estimate (Eq.~\eqref{eq:randomized_smoothing}) or the central estimate (Eq.~\eqref{eq:randomized_smoothing_central}) as a gradient estimate requires at most $n$ times more iterations than the standard gradient method~\citep{nesterov2017random}.

To summarize, although the assumptions in \cite{spall2005introduction, wright2015coordinate, nesterov2017random} vary, the underlying ideas are of the same nature; sampling along random directions may match the performance of gradient descent while requiring $n$ times fewer function evaluations.

\begin{remark}
    Importantly, these results do not take into account the possibility of making computations in parallel (see Section \ref{section:parralel}). In practice, performing function evaluation in parallel offers the possibility of taking Monte Carlo averages with multiple samples at no extra computational cost. 
\end{remark}

\subsection{Simulated Annealing}

So far, we have seen that performing random approximate gradient descent can be sample efficient in the convex setting. However, one may ask whether these algorithms can converge to global solutions in nonconvex settings, as it is the case for the Greedy Local Search introduced in Algorithm~\ref{algo:local_search}. 
It is well known that, in its vanilla form, gradient descent is prone to local minima. However, additive Gaussian noise  can lead to asymptotic global convergence guarantees. To understand why, we need to study the following Langevin dynamics:
\begin{align}
    dX_t &= \nabla_x \log p(X_t) + dW_t.
\end{align}
Here $p$ is a probability distribution, and $W_t$ is a Brownian motion. $\nabla_x \log p$ is the score function of the distribution. Note that the gradient is taken with respect to the input of $p$ (not with respect to the parameters of $p$ as in the log-likelihood trick of Gaussian smoothing). This Stochastic Differential Equation (SDE) has been extensively studied, and it can be shown that $X_t$ converges to the distribution defined by $p$~\citep{roberts1996exponential}. Consequently, this SDE can be used to sample from a distribution with probability density~$p$. One way to leverage this result in the context of global optimization is to choose the following probability distribution:
\begin{align}
    p \propto  \exp \left(-\frac{1}{\lambda} f \right), \label{eq:exp_energy_eq}
\end{align}
where $\lambda > 0$ is called the temperature. As $\lambda \rightarrow 0 $, the distribution concentrates on the set of global minima of $f$~\citep{andrieu2003introduction}. 
Furthermore, the score function of this distribution is proportional to the gradient of $f$:
\begin{align}
    \nabla_x \log p(x) = - \frac{1}{\lambda} \nabla_x   f(x). \label{eq:energy_eq}
\end{align}
Then, this SDE can be discretized to derive a recursive algorithm.  
Following this idea, \cite{gelfand1991recursive} show that with an appropriate schedule of step sizes $\alpha_k$ and $\gamma_k$, the discretized Stochastic Langevin Gradient Dynamics,
\begin{align}
    x_{k+1} = x_k - \alpha_k  g_k + \gamma_k \epsilon_k, \label{eq:slgd}
\end{align}
guarantees asymptotic convergence to global solutions. Here, $\epsilon_k$ is Gaussian noise and $g_k$ is of the form ${(\nabla f(x_k) + \xi_k)}$, with $\xi_k$ representing stochastic noise. Hence, $g_k$ can be interpreted as a noisy approximation of the gradient.
Intuitively, the Gaussian noise $\epsilon_k$ helps to escape local minima.
The step sizes $\alpha_k$ and $\gamma_k$ are chosen such that the corresponding SDE has a temperature $\lambda$ that tends slowly to zero. This is called Simulated Annealing. Intuitively, slowly decreasing the temperature guides the iterates towards a global solution.

These results can be used to prove the global convergence of SPSA. More precisely, the iterative scheme in Equation~\eqref{eq:slgd} with $g$ as the SPSA estimate defined in Equation~\eqref{eq:SPSA} converges to a global minimum~\citep{maryak2001global}.
In fact, in its default version (i.e. without additive noise), SPSA converges to a global minimum~\citep{maryak2001global}. This is because the noise of the estimate (i.e. $\xi_k$) is enough to escape local minima and achieve global convergence. We are not aware of such results in the context of randomized smoothing. However, we can conjecture that similar results may be valid.

To summarize, we have seen that stochasticity can help gradient descent algorithms reach global solutions. Although these results are highly relevant theoretically, they have limited practical impact, as they are only asymptotic. 

\begin{remark}
    At first glance, it might be hard to relate the discretized Langevin update to the Greedy Local Search in Algorithm~\ref{algo:local_search}. However, following the observation from \cite{toussaint2024nlp}, the \emph{if} statement ensuring a strict decrease can be related to the Metropolis-Hastings algorithm~\citep{andrieu2003introduction}. We refer the reader to Appendix~\ref{appendix:GLSandMALA} for more details.
\end{remark}

\section{Trajectory Optimization}\label{section:TO}

In this section, we show how random search allows us to understand derivative-free optimization algorithms that are commonly used in robotics. In practice, Trajectory Optimization aims to solve
\begin{align}\label{eq:TO}
   \min_{u_0, u_1, \dots, u_{T-1}}& \; \sum_{t=0}^{T-1} c_t(x_t, u_t) +c_T(x_T),  \\
    \mbox{such that }& \, x_0 = x, \,\mbox{and} \;\;  x_{t+1} = f_{\text{dyn}}(x_t, u_t). \nonumber
\end{align}
Here $x$ denotes the state, $u$ the control variable, $(c_t)_t$ the running cost functions, $c_T$ the terminal cost function, $T$ the time horizon, and $f_{\text{dyn}}$ the dynamics. Note that we abuse the notation by using $x$ to denote the state instead of the optimization variable. 
Importantly, the constraints are implicit; the optimization is performed only with respect to the control variables. Therefore, this problem is unconstrained and is referred to as the single shooting approach. Ultimately, it can be written in the form of Problem~\eqref{eq:main_problem}.
The dimension of the search space is $n = T n_u$, where $n_u$ denotes the dimension of the control inputs.

In practice, directly optimizing over all control inputs can lead to very high-dimensional search spaces, especially for long horizons. A common approach is therefore to parameterize the control trajectory using splines~\citep{de1978practical}. Typical choices include zero-order hold, piecewise-linear interpolation, and cubic Hermite splines~\citep{howell2022predictive}. 

\subsection{Predictive sampling}

In robotics, Predictive sampling~\citep{howell2022predictive} is one of the simplest random search approaches. This method can be seen as a variation of the Greedy Local Search (Algorithm~\ref{algo:local_search}). Instead of sampling one search direction, it samples $K$ search directions and keeps the best. If no direction provides an improvement, the previous estimate is kept.
Algorithm~\ref{algo:predictive_sampling} summarizes the procedure.
\begin{algorithm}[!ht]
\DontPrintSemicolon
\KwInput{$x_0 \in \mathbb R^n$, K}
$ x \gets x_0$\;
\While{stopping criterion is not met}{
Sample $d_1, \dots, d_K$ \tcp{with independent centered Gaussian distributions} 
$D \gets \{ x \}$ \;
\For{$k=1, \dots K$}{ Add $x+d_k$ to $D$}
$x \gets \operatorname{argmin}_{\tilde{x} \in D} f(\tilde{x})$
}
\KwOutput{$x$}
\caption{Predictive Sampling}\label{algo:predictive_sampling}
\end{algorithm}

Importantly, all the function evaluations can be performed in parallel.
While this approach can be quite effective, recent sampling-based MPC demonstrations typically rely on more sophisticated algorithms~\citep{li2024drop, xue2024full}.

\subsection{The log-sum-exp transform: MPPI}

Model Predictive Path Integral (MPPI)~\citep{williams2016aggressive} is a derivative-free TO method initially derived from an information-theoretic perspective. MPPI iteratively samples around the current guess and updates it using a weighted exponential average. More specifically, given a current guess~$x$, MPPI samples $K$ variables $x_k$ following independent Gaussian distributions centered in $x$ (i.e. ${x_k \sim \mathcal{N}(x, \Sigma)}$). Then, it performs an update according to the following rule:
\begin{align}
    x \gets (1 - \alpha) x + \alpha \sum_{k=1}^K w_k x_k, \label{eq:update_rule_MPPI}
\end{align}
where
\begin{align} \label{eq:MPPI_weigth}
    w_k = \frac{ \exp (- \frac{1}{\lambda} (f(x_k) - \rho) )}{\sum_{j=1}^K  \exp (- \frac{1}{\lambda} (f(x_j) - \rho) ) },
\end{align}
where $\rho = \min_j f(x_j)$ (to prevent numerical stability issues), and where $\alpha$ is the step-size (or learning rate). These weights, $w_k$, are called exponential average weights. Algorithm~\ref{algo:MPPI} describes the complete procedure. Note that originally, MPPI was introduced as an MPC method with a single update per control step, i.e., as a real-time iteration MPC scheme~\citep{williams2016aggressive}. However, in this work, we interpret MPPI as an optimization algorithm and use the term MPPI to denote the iterative procedure defined by the update rule~\eqref{eq:update_rule_MPPI}.

\begin{algorithm}[!ht]
\DontPrintSemicolon
\KwInput{$x_0 \in \mathbb R^n, \Sigma, K, \alpha$}
$ x \gets x_0 $\;
\While{stopping criterion is not met}{
    Sample $x_{1:K} \sim \mathcal{N}(x, \Sigma)$ \tcp*{sample}  
    Compute weights $w_1, \dots w_K$\; 
    $x \gets (1-\alpha) x + \alpha \sum\limits_{k=1}^{K} w_k x_k$ \tcp*{update}  
}
\KwOutput{$x$}
\caption{MPPI (with learning rate)}
\label{algo:MPPI}
\end{algorithm}

Let's see how this update rule can be interpreted as a variation of Gaussian smoothing. Instead of the surrogate function defined in Equation~\eqref{eq:surrogate-rs}, let's consider the (continuous) log-sum-exp transform function.
\begin{align}
f_{\mu,\lambda}(x) = - \lambda \log \left( \mathbb  E \left[ \exp\left(- \frac{1}{\lambda} f(x + \mu \epsilon) \right) \right] \right),~\label{eq:mppi_surrogate}
\end{align}
where $\epsilon \sim \mathcal{N}(0, \Sigma)$ and where $\lambda > 0$ is a scalar called the temperature. As $\mu$ goes to zero, we recover $f$. Therefore, the gradients of this surrogate can be used to approximate the gradients of the original function, $f$. Furthermore, as $\lambda \rightarrow \infty, f_{\mu,\lambda} \rightarrow f_{\mu}$~\citep{scaman2020simple}. Consequently, this surrogate can be seen as a generalization of Gaussian smoothing.
The gradient of the surrogate reads:
\begin{align}
\hspace{-0.1cm}  \nabla f_{\mu, \lambda}(x) = \frac{ - \lambda \mathbb E \left[  \exp\left(- \frac{1}{\lambda} f(x + \mu \epsilon) \right) \Sigma^{-1}  \epsilon \right]}{ \mu \mathbb   E \left[ \exp\left(- \frac{1}{\lambda} f(x + \mu \epsilon) \right) \right]  }.    ~\label{eq:mppi_surrogate_grad}
\end{align}

 We now show that MPPI approximates a natural gradient step~\citep{amari1998natural}. More precisely, MPPI approximates the following update:
\begin{align}
x \gets   x - \frac{\alpha}{\lambda} F^{-1}   \nabla f_{\mu, \lambda}(x), 
\end{align}
with $\mu=1$ and where $F$ is the Fisher Information Matrix (FIM), which is equal to the inverse of the covariance matrix for Gaussian distributions. The idea of the natural gradient is to build an update step that is invariant to changes of variables. More details on the natural gradient are provided in Appendix~\ref{appendix:natural_gradient}. 
\begin{theorem}
With $\mu = 1$, we have:
\begin{align} \label{eq:update_analytical_MPPI}
 x - \frac{\alpha}{\lambda} F^{-1}   \nabla f_{\mu, \lambda}(x) = (1 - \alpha) x + \alpha \mathbb E_{z\sim \mathcal{N}(x, \Sigma)} \left[  w_f(z) z \right]
\end{align}
   where $F=\Sigma^{-1}$ is the Fisher information matrix, and where the weights are defined as:
\begin{align}\label{eq:continuous_mppi_weights}
  w_f(z) =  \frac{   \exp\left(- \frac{1}{\lambda} f(z ) \right)  )  }{ \mathbb   E_{z\sim \mathcal{N}(x, \Sigma)}  \left[ \exp\left(- \frac{1}{\lambda} f(z) \right) \right]  }. 
\end{align}
\end{theorem}
Intuitively, $w_f(z)$ is the continuous version of the exponential average weights.
In practice, one can use Monte Carlo approximation to estimate both expectations. Let's denote $x_{1:K}$, $K$ independent samples following the distribution $\mathcal{N}(x, \Sigma)$.  We have
\begin{align}
 x - \frac{\alpha}{\lambda} F^{-1}   \nabla f_{\mu, \lambda}(x) &\approx   (1-\alpha)x + \alpha \sum\limits_{k=1}^K w_k x_k.
\end{align}
Hence, we recover the MPPI update. Note that, similarly to the randomized smoothing case, this update is invariant to additive translation of the function $f$. 
\begin{remark}
   Note that \cite{wagener2019online} also uses the natural gradient to connect MPPI to mirror descent. In the Appendix~\ref{appendix:natural_gradient}, we explain how this relates to the above derivations. 
\end{remark}

Interestingly, if $\Sigma=\sigma^2 I$, then $ \frac{\sigma^2}{\lambda} \leq \frac{1}{L}$, where $L$ is the Lipschitz function of the surrogate function~\citep{scaman2020simple}. Therefore, the step size chosen by MPPI with $\alpha = 1$ can be interpreted as a conservative estimate of the standard optimal step size from convex optimization~\citep{nesterov2018lectures}.

\begin{figure}[ht!]
    \centering
    \includegraphics[width=\linewidth]{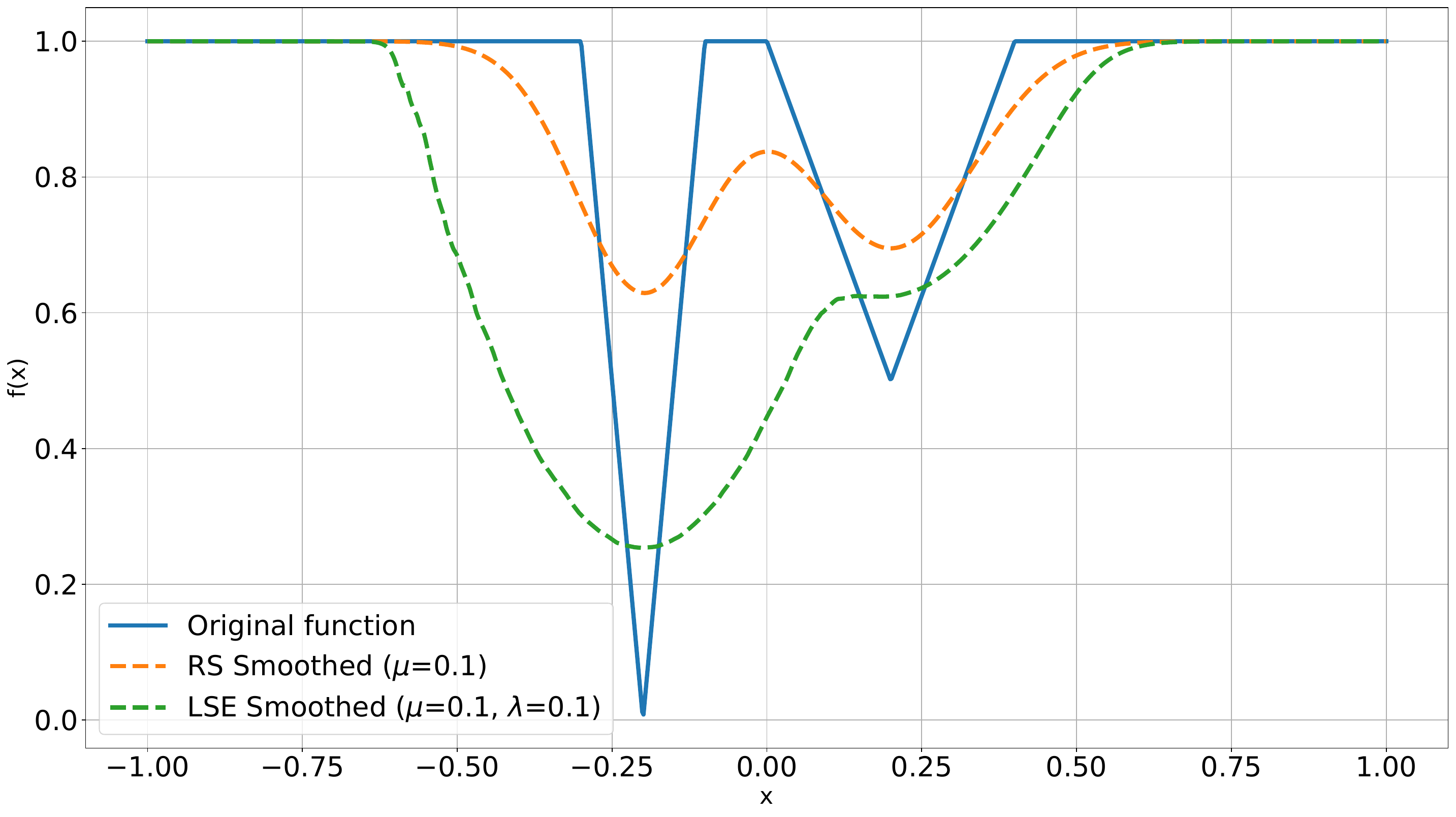}
    \caption{Randomized smoothing (RS) compared to log-sum-exp (LSE) smoothing (with $\Sigma=I$).}
    \label{fig:lse-illustration}
\end{figure}

Figure~\ref{fig:lse-illustration} illustrates how the log-sum-exp smoothing differs from Nesterov's randomized smoothing. The expectations are estimated with $100,000$ samples. This simple example function has two types of local minima: one minimum is global but narrow, and the other is non-global but wide. Interestingly, Gaussian smoothing gives similar importance to both minima; the two minima now have similar depths and widths. In contrast, the log-sum-exp smoothing gives more importance to the global minimum; the global minimum is much wider and almost absorbs the local one. 

To the best of our knowledge, \cite{scaman2020simple} propose the first explanation of the success of MPPI from an optimization perspective. \cite{scaman2020simple} shows that the log-sum-exp smoothing can be interpreted as an interpolation between the default randomized smoothing and the Moreau envelope. The idea is that the Moreau envelope is a better approximation than RS but is harder to compute. MPPI elegantly trades off approximation quality and computational complexity. At the cost of more samples, relying on this surrogate can improve convergence rates~\citep{scaman2020simple}.
Intuitively, this estimator requires many samples because of the presence of the expectation in the denominator in Equation~\eqref{eq:mppi_surrogate_grad}. 
With the progress of parallel computing, this requirement for multiple samples might not be a limitation at all. However, using many samples renders the update step essentially deterministic, which makes the algorithm prone to getting trapped in local minima.

\begin{remark}
The log-sum-exp smoothing resembles the risk-seeking control formulation~\citep{whittle1981risk}. We refer the reader to Appendix~\ref{appendix:risk} for a discussion on the topic.
\end{remark}

\begin{remark}
The log-sum-exp smoothing is reminiscent of the simulated annealing idea. Similarly to simulated annealing, as the temperature $\lambda$ approaches $0$, the surrogate concentrates around the global minima. In fact,
\begin{align*}
 p \propto \exp \left(-\frac{1}{\lambda}f_{\mu, \lambda} \right) =  \mathbb  E \left[ \exp\left(- \frac{1}{\lambda} f(x + \mu \epsilon) \right) \right] .
\end{align*}
Therefore, applying the Langevin update to the log-sum-exp surrogate function amounts to applying Gaussian smoothing to the probability distribution proportional to $e^{-\frac{1}{\lambda}f}$.
\end{remark}

\subsection{Covariance Matrix Adaptation (CMA)}~\label{sec:CMA}

So far, we have kept the covariance matrix $\Sigma$ fixed across various iterations. However, in practice, it is not necessarily clear how to choose this parameter according to the problem. In this section, we show how to derive a Covariance Matrix Adaptation (CMA) scheme.
Interestingly, the optimization of the smooth surrogate function can be interpreted as an optimization in the space of Gaussian probability distributions. Indeed, optimizing Equation~\eqref{eq:surrogate-rs} can be interpreted as a search over the mean of a Gaussian distribution. More specifically, we can write:
\begin{align*}
\mathbb E_{\epsilon\sim \mathcal{N}(0, \Sigma)} [ f(x + \epsilon )] = \mathbb E_{z\sim \mathcal{N}(x,  \Sigma)} \left[ f(z) \right].
\end{align*}
Therefore, Gaussian smoothing can be seen as the search for a Gaussian distribution that minimizes the expectation of $f$ under that distribution. This distribution can be interpreted as the belief of where the global minimum of the function $f$ might be. A natural generalization is to not only optimize the mean but also the covariance. More specifically, one can apply gradient descent to the following problem:
\begin{align}
    \min_{\theta }  J(\theta)\,
\mbox{ where }&
\,     J(\theta) = \mathbb E_{z\sim \mathcal{N}(x, \Sigma)} \left[ f(z) \right], \nonumber\\
\mbox{ and } &    \theta = \begin{bmatrix}
        x\\
        \operatorname{vec}\left( \Sigma \right)
    \end{bmatrix}.
\end{align}
where the $\operatorname{vec}$ operator transforms a matrix into a vector by concatenating its columns (i.e., for a matrix ${A = [a_1, \dots, a_d]}$,
$ \operatorname{vec}(A) = [a_1^T, \dots, a_d^T]^T $).
Importantly, $J(\theta)$ is minimal when the distribution $\mathcal{N}(x, \Sigma)$ concentrates around the minima of $f$~\citep{ollivier2017information}. Note that this idea does not necessarily need to rely on Gaussian distributions. Similarly to the randomized smoothing case, we can use the log-likelihood trick:
\begin{align}
\nabla_{\theta} J(\theta) =  \mathbb E_{z\sim \mathcal{N}(x, \Sigma)} \left[ f(z) \nabla_{\theta} \log (p_{\theta}(z))\right]. \label{eq:loglikelihood_trick}
\end{align}
Here, the gradient of the probability distribution is taken with respect to both the mean and the covariance parameters. In order to derive an update invariant to the parameterization of the Gaussian distribution, we aim to study the natural gradient update, ${\Delta \theta =  F(\theta)^{-1}\nabla_{\theta} J(\theta)}$.
\begin{theorem} \label{theorem:akimoto}
\cite{akimoto2010bidirectional} shows that the following property holds:
\begin{align*}
   F(\theta)^{-1} \nabla_{\theta} \log (p_{\theta}(z)) = \begin{bmatrix}z - x \\
      \operatorname{vec}\left( \Sigma  \left(z - x  \right) \left(z- x  \right)^\top - \Sigma\right)
   \end{bmatrix} 
\end{align*}
\end{theorem}

Using this theorem and the linearity of the expectation operator, we find that $\Delta \theta$ is equivalent to:
\begin{align}\label{eq:CMA_in_expectation}
\Delta \Sigma &= \mathbb E_{z\sim \mathcal{N}(x, \Sigma)} \left[f(z) (\left(z - x  \right) \left(z- x  \right)^\top - \Sigma )\right], \nonumber  \\
\Delta x &=  \mathbb E_{z\sim \mathcal{N}(x, \Sigma)} \left[ f(z) (z -  x)  \right] .
\end{align}
Similarly to the randomized smoothing case, this update step averages quantities that rely solely on function evaluations of the original function. Following the natural gradient is especially relevant in this setting, as it is crucial that the update does not depend on the choice of parameterization of the Gaussian distribution.
We refer the reader to Appendix~\ref{appendix:natural_gradient} for more details on the natural gradient.
In practice, this update step can be approximated via Monte Carlo sampling. Given samples $x_k\sim \mathcal{N}(x, \Sigma)$, and denoting $w_k = f(x_k)$, the approximate update reads
\begin{align}\label{eq:CMA_update}
\Sigma &\gets (1-\alpha_{m} \sum\limits_{k=1}^{K} w_k) \Sigma + \alpha_{m} \sum\limits_{k=1}^{K} w_k (x_k - x)(x_k - x)^\top, \nonumber\\
x &\gets (1-\alpha_{\Sigma} \sum\limits_{k=1}^{K} w_k) x + \alpha_{\Sigma} \sum\limits_{k=1}^{K} w_k x_k .
\end{align}
where $(\alpha_{m},\alpha_{\Sigma})$ are the learning rates for the mean and covariance. We leave the possibility of choosing different learning rates for the mean and covariance, as we will see in the experiments that this can greatly improve performance.

This allows us to recover the CMA scheme (see Algorithm~\ref{algo:CMA}). Note that it is important to update the covariance matrix first, as the covariance update must rely on the old $x$ (i.e. the one used to sample). 
As with Gaussian smoothing, a constant can be subtracted from $f$ without changing the expectation in Equation~\eqref{eq:loglikelihood_trick}. Therefore, another choice of weights could be $w_k = f(x_k) - f(x)$, as this renders the update invariant to translations of $f$.

\begin{algorithm}[!ht]
\DontPrintSemicolon
\KwInput{$x_0 \in \mathbb R^n, \Sigma_0, K, \alpha_{m}, \alpha_{\Sigma}$}
$ (x, \Sigma) \gets (x_0, \Sigma_0)$\;
\While{stopping criterion is not met}{
    Sample $x_{1:K} \sim \mathcal{N}(x, \Sigma)$ \;
    Compute weights $w_1, \dots w_K$\; 
    $\Sigma \gets (1-\alpha_{\Sigma} \sum\limits_{k=1}^{K} w_k) \Sigma $  
$\phantom{.......}  + \,\, \alpha_{\Sigma} \sum\limits_{k=1}^{K} w_k (x_k - x)(x_k - x)^\top$     
    \;
    $x \gets (1-\alpha_{m} \sum\limits_{k=1}^{K} w_k) x + \alpha_{m} \sum\limits_{k=1}^{K} w_k x_k$ 
}
\KwOutput{$x$}
\caption{Covariance Matrix Adaptation (CMA)}
\label{algo:CMA}
\end{algorithm}

An important point that we overlooked is whether the covariance update rule in Equation~\eqref{eq:CMA_update} maintains the matrix positive definite. In fact, if $ 0\leq \alpha_m, \alpha_{\Sigma}  < 1$, $w_i\geq 0$, and the weights sum to one, then the covariance matrix is always positive definite, provided that we start with a positive definite matrix~\citep{akimoto2010bidirectional}. 
However, in practice, there is no reason that the function $f$ would be such that the weights satisfy this property. 

One possible solution is to design an update that operates on exponential coordinates so that the covariance matrix naturally stays positive definite. This is the idea behind xNES~\citep{wierstra2014natural}.
Thanks to the invariant properties of the natural gradient, both update rules align when the step size approaches zero.

Another approach could be to use the log-sum-exp transformation used for MPPI. Indeed, this transformation naturally maintains the weights as positive and ensures that they sum to one. In fact, the following theorem holds:
\begin{theorem}
Given $\theta = \begin{bmatrix}
        x\\
        \operatorname{vec}\left( \Sigma \right)
    \end{bmatrix}$ and the log-sum-exp cost, 
\begin{align}\label{eq:lse_cost_4_cma_theorem}
    J_{\lambda}(\theta) = - \lambda \log \left( \mathbb  E_{z\sim \mathcal{N}(x, \Sigma)} \left[ \exp\left(- \frac{1}{\lambda} f(z) \right) \right] \right)
\end{align}
the natural gradient update  ${\Delta \theta = \lambda^{-1}  F(\theta)^{-1}\nabla_{\theta} J_{\lambda}(\theta)}$ reads
\begin{align}\label{eq:MPPI-CMA_in_expectation}
\Delta \Sigma &= \mathbb E_{z\sim \mathcal{N}(x, \Sigma)} \left[w_f(z) (\left(z - x  \right) \left(z- x  \right)^\top - \Sigma )\right], \nonumber  \\
\Delta x &=  \mathbb E_{z\sim \mathcal{N}(x, \Sigma)} \left[ w_f(z) (z -  x)  \right] .
\end{align}
where $w_f(z) $ is defined as in \eqref{eq:continuous_mppi_weights}.
\end{theorem}

The proof is straightforward and is based on Theorem~\ref{theorem:akimoto} and the following equality:
\begin{align*}
    F(\theta)^{-1}\nabla J_{\lambda}(\theta) = - \lambda \mathbb E \left[  w_f(z) F(\theta)^{-1} \nabla_{\theta} \log (p_{\theta}(z))  \right]
\end{align*}

As for the previous algorithms, the update~\eqref{eq:MPPI-CMA_in_expectation} can be approximated via Monte Carlo sampling by replacing each expectation with finite sums. We refer to this algorithm as MPPI-CMA. Clearly, this algorithm is an instance of Algorithm~\ref{algo:CMA} with the exponential average weights from the MPPI update~\eqref{eq:MPPI_weigth}. Therefore, with $\alpha_{\Sigma} = 0$, MPPI-CMA reduces to MPPI. 

As expected, because the exponential average weights~\eqref{eq:MPPI_weigth} are positive and sum to one, MPPI-CMA maintains its covariance matrix positive definite. Also, the exponential average weights make MPPI-CMA invariant to additive translation of the objective. However, one could go further and design weights invariant to any monotone transformation of the objective by choosing arbitrary weights $w_k$ sorted so that their order matches that of the function values  $f(x_k)$.
As pointed out by \cite{akimoto2010bidirectional, wierstra2014natural, ollivier2017information}, this recovers the CMA-ES algorithm (without evolution path)~\citep{hansen2001completely}.

\begin{remark}\label{rmk:CEM}
Another choice of weight to maintain invariance to any monotone transform could be to assign $w_k = \frac{1}{K_e}$ to the best $K_e$ samples and $0$ to the others (by best, we mean samples with the lowest value $f(x_k)$). As pointed out by \cite{ollivier2017information}, this elitist weighting provides an interesting connection to the well-known Cross-Entropy Method (CEM), a popular algorithm that can be used for optimization. The only difference with the original CEM is that the natural gradient update computes the covariance update using the old $x$ (i.e. the one used to sample) instead of the new one~\citep{ollivier2017information}.  
For reference, we provide in the Appendix~\ref{appendix:CEM} the main steps of the CEM algorithm. 
\end{remark}

Table~\ref{tab:algorithms} summarizes all the possible variations of the CMA algorithm, depending on the choice of weights.
We refer the reader to the work of \cite{ollivier2017information} for a formal justification of the weights of CEM and CMA-ES.

\begin{table}[!ht]
\centering
\begin{tabular}{|l|c|c|}
\hline
 &   Weights \\
\hline
 MPPI-CMA  & Exponential Average (Eq. \ref{eq:MPPI_weigth})\\
\hline
  CMA-ES$^{\star}$ & Ordering~\citep{hansen2001completely}  \\
\hline
 CEM$^{\star\star}$& Elitist (Remark~\ref{rmk:CEM}) \\
\hline
\end{tabular}
\caption{Variations of Algorithm~\ref{algo:CMA}. $^{\star}$We refer to CMA-ES with no evolutionary path. $^{\star\star}$Note that CEM updates the mean first.}
\label{tab:algorithms}
\end{table}

So far, the algorithms we have covered are generic. That is, they can be applied to any optimization problem and are not specific to TO. Next, we will see how the CMA scheme can exploit the specific structure of TO.

\subsection{Block-diagonal CMA}

When dealing with TO, as in Problem~\eqref{eq:TO}, it is often preferred to design algorithms that can achieve linear complexity with respect to the time horizon $T$. This is typically achieved by sampling the control inputs from each time step independently. In other words, one can use a block-diagonal covariance matrix (with $T$ blocks, one for each control input). 
Clearly, this makes the update rules of Predictive Sampling and MPPI easily implementable with linear complexity with respect to the time horizon.

However, this is less straightforward in the case of the CMA update. Specifically, the update~\eqref{eq:MPPI-CMA_in_expectation} has no reason to maintain the block-diagonal structure of the covariance matrix. This is because $\Sigma$ is not the appropriate representation to work with.
Instead of applying the natural gradient descent to the full matrix $\Sigma$, one can apply it to each block.
To be more precise, let $x = (x_1^T, \dots, x_T^T)^T$ be a trajectory, i.e., the concatenation of $T$ vectors of dimension $d$ (which could describe control inputs with splines, for example), and let $\Sigma = \operatorname{Diag}(\Sigma_1, \dots, \Sigma_T)$ be a block-diagonal matrix with $T$ blocks of size $d\times d$. We can show the following theorem.

\begin{theorem}\label{th:block-diagonal}
Given the log-sum-exp cost~\eqref{eq:lse_cost_4_cma_theorem} and
\begin{align}\label{eq:theta_for_block}
    \theta =
\begin{bmatrix}
    x_1\\ \vdots\\x_T\\ \operatorname{vec} \left( \Sigma_1\right)\\ \vdots\\ \operatorname{vec} \left( \Sigma_T\right)
\end{bmatrix}
\end{align}
the natural gradient update  ${\Delta \theta =  \lambda^{-1} F(\theta)^{-1}\nabla_{\theta} J_{\lambda}(\theta)}$ reads: 
\begin{align} 
\Delta \Sigma_t &= \mathbb E_{z\sim \mathcal{N}(x, \Sigma)} \left[w_f(z) (\left(z_t - x_t  \right) \left(z_t- x_t  \right)^\top - \Sigma )\right], \nonumber  \\
\Delta x_t &=  \mathbb E_{z\sim \mathcal{N}(x, \Sigma)} \left[ w_f(z) (z_t -  x_t)  \right] .
\end{align}
for $t=1, \dots, T$, and where $w_f(z) $ is defined as in \eqref{eq:continuous_mppi_weights}.
\end{theorem}
In other words, one can apply the CMA update independently to each block. The only shared information is the weight computation that relies on function evaluations based on the entire trajectory. Indeed, the samples $z$ are full trajectories. A proof is provided in the Appendix~\ref{appendix:block}. As before, the expectations can be approximated with Monte Carlo sampling, leading to the expectations being replaced by finite sums and the continuous weight, $w_f(z)$, being replaced by the MPPI weight.
We refer to this approach as block-diagonal CMA.
Importantly, this formulation can be implemented sequentially, ensuring linear computational complexity with respect to the time horizon~$T$. This reduced complexity can be especially relevant for MPC applications where computations need to be real-time.
\begin{remark}
MPPI with block-diagonal CMA is the same as $\text{PI}^2$-CMA without temporal averaging~\citep{stulp2012path}. This method was also used by~\cite{bhardwaj2022storm}.
\end{remark}
\begin{figure*}
\centering
\begin{subfigure}{0.49\linewidth}\centering
  \includegraphics[width=\linewidth]{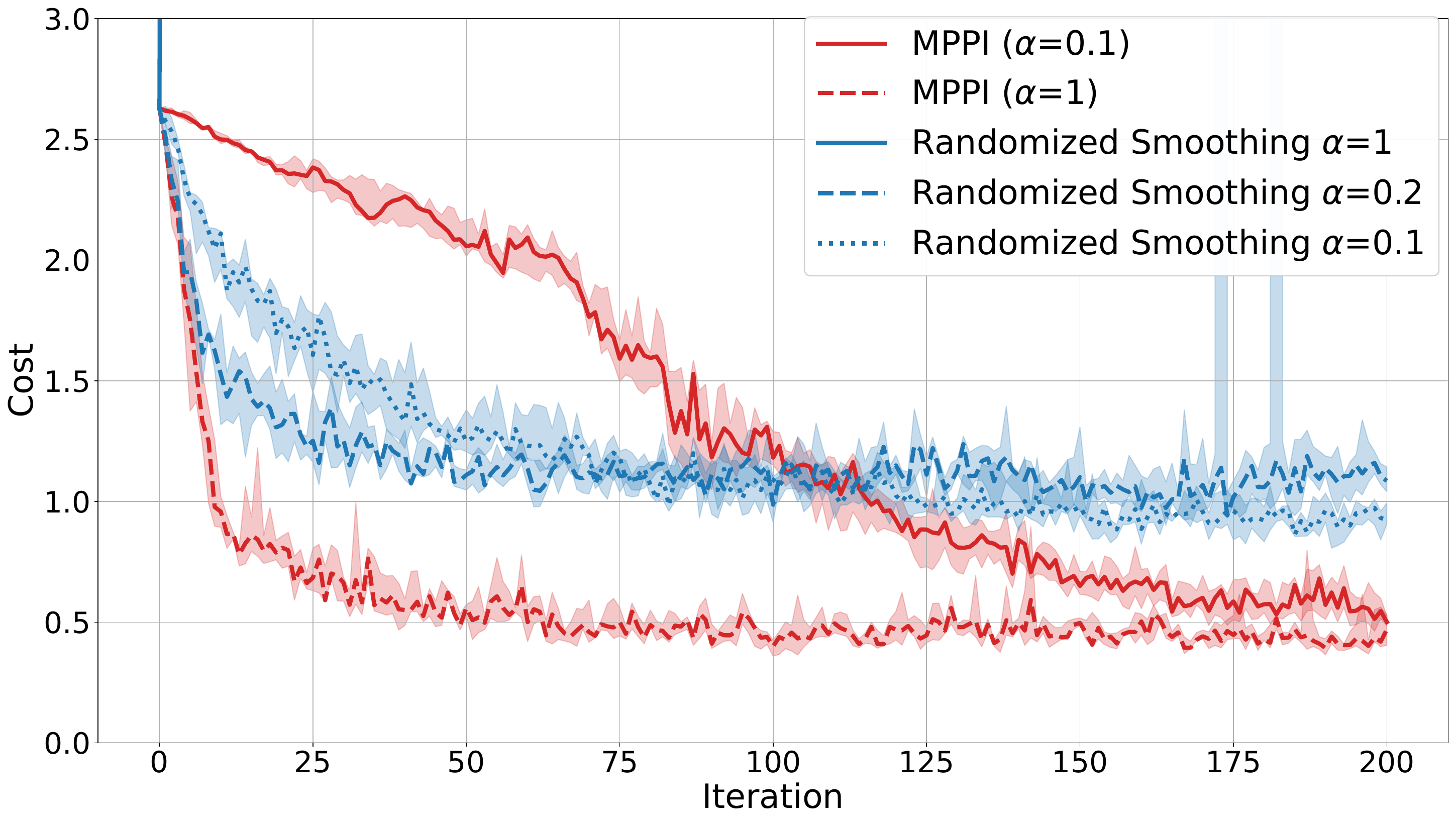}
  \caption{Go2Walk}\end{subfigure}\hfill
\begin{subfigure}{0.49\linewidth}\centering
  \includegraphics[width=\linewidth]{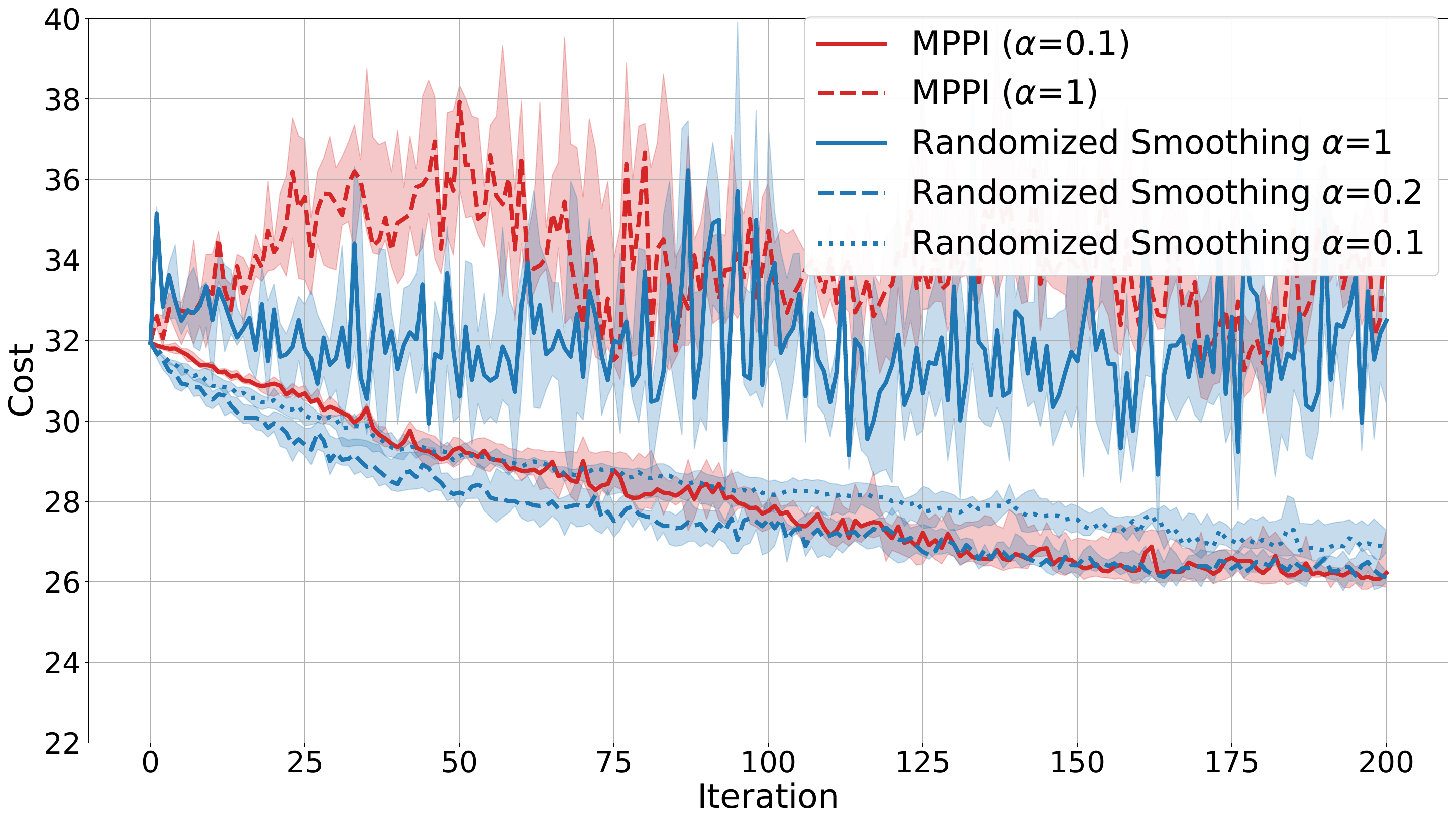}
  \caption{HumanoidStandup}\end{subfigure}
\caption{Cost according to the number of iterations for different test problems. MPPI outperform RS but needs a task-dependent learning-rate.}
\label{fig:rs-mppi}
\end{figure*}
\begin{figure*}
\centering
\begin{subfigure}{0.49\linewidth}\centering
  \includegraphics[width=\linewidth]{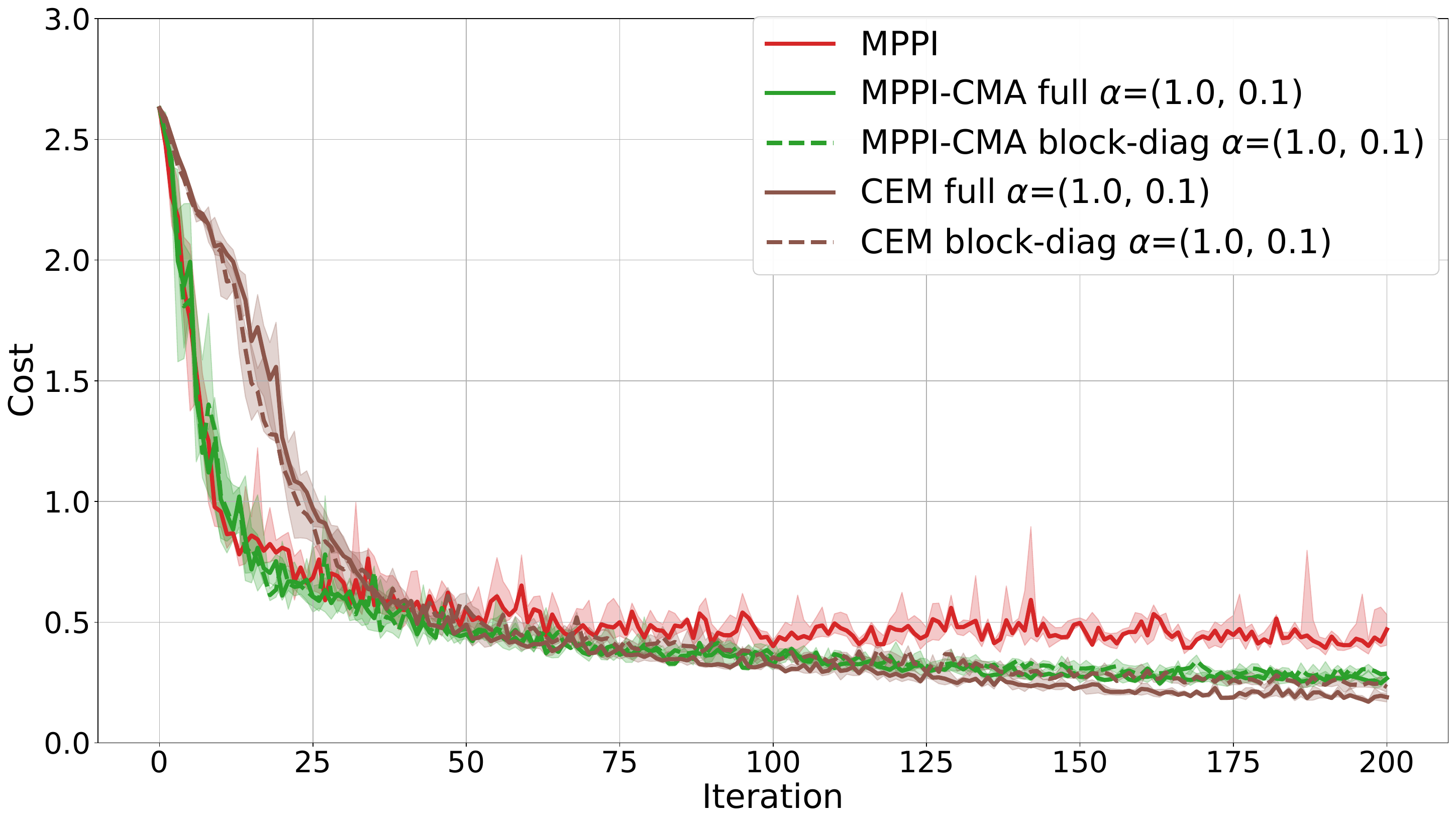}
  \caption{Go2Walk}\end{subfigure}\hfill
\begin{subfigure}{0.49\linewidth}\centering
  \includegraphics[width=\linewidth]{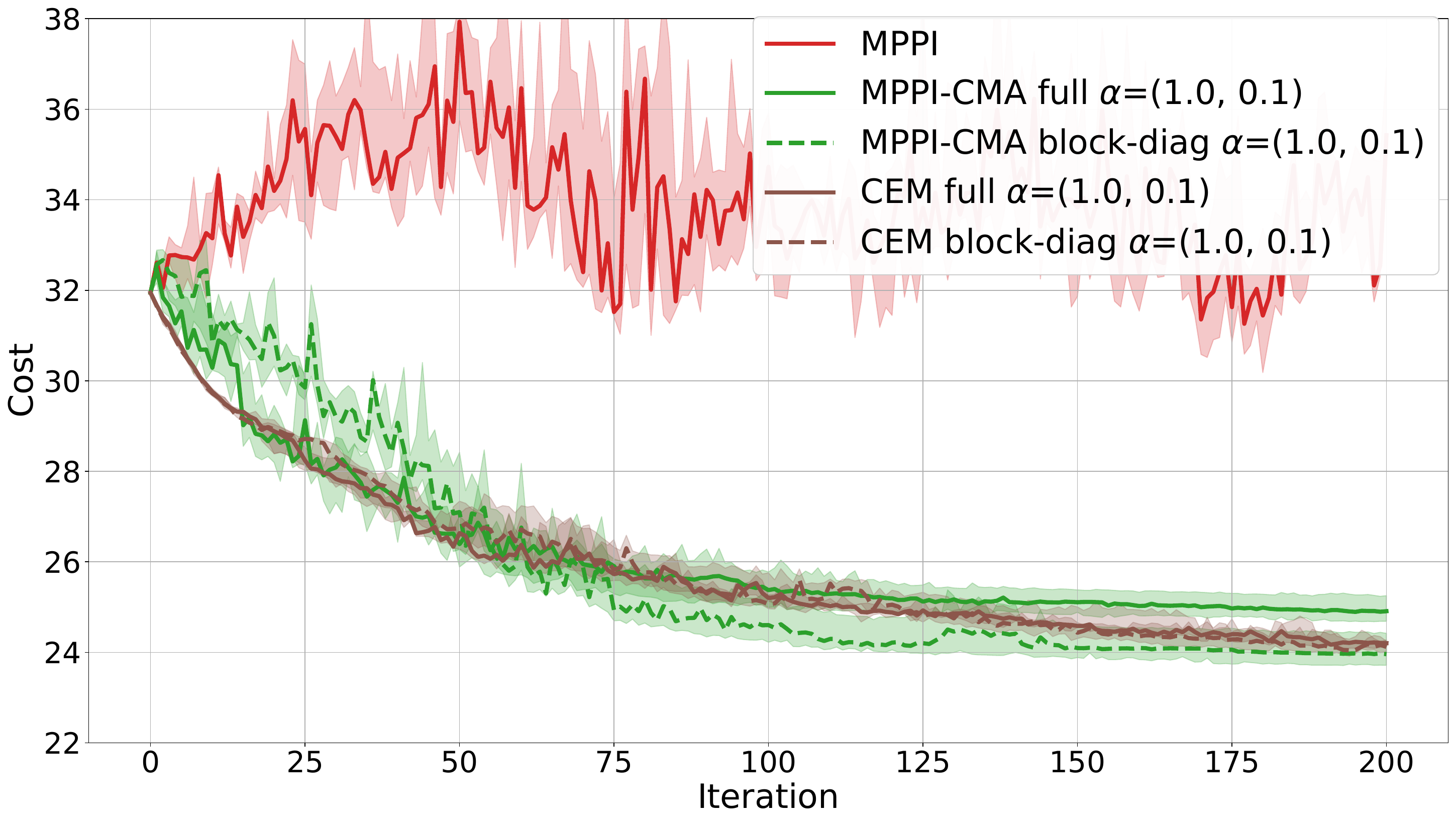}
  \caption{HumanoidStandup}\end{subfigure}
\caption{Cost according to the number of iterations for different test problems. The CMA update allows to use the same learning rate parameters across multiple tass. The full and block-diagonal covariance update perform similarly.}
\label{fig:cov}
\end{figure*}

\subsection{Numerical experiments}

We introduce six TO benchmark problems based on Hydrax~\citep{kurtz2024hydrax}: Cartpole, DoubleCartPole, PushT, HumanoidBalance, HumanoidStandup, and Go2Walk. Control bounds are enforced with a penalty term in the cost function.
All results are averaged over six seeds.
More details about the cost functions and additional visualizations are available online~\footnote{\url{https://github.com/ajordana/zoo-rob}}.
We compare the performance of Predictive Sampling, randomized smoothing (RS), CEM, CMA-ES, MPPI, and MPPI-CMA. All algorithms use $2048$ samples per iteration. For CEM and MPPI-CMA. we implement both the full and the block-diagonal versions of CMA.
For MPPI and MPPI-CMA, we use a temperature of $\lambda=0.1$.
As the weights of RS are not bounded, we prune out rollouts with very large costs. Otherwise, RS systematically diverges (even for very small learning rates). This illustrates another strength of MPPI, CMA-ES, or CEM: when a rollout creates a very large cost, the corresponding weight will be equal to or almost equal to zero, and therefore will not bring numerical instability.
We present three key findings:


First, MPPI outperforms RS. Figure~\ref{fig:rs-mppi} shows the performance of MPPI and RS with various learning rates on two challenging problems. MPPI performs better; however, we can see that while a learning rate of $0.1$ performs better than a learning rate of $1$ on the Go2Walk example, it is the opposite on the HumanoidStandup task. This illustrates the task dependent tuning of the learning rate that MPPI requires.

Second, the CMA update improves performance and allows for a set of parameters that work well on multiple tasks (as opposed to vanilla MPPI, where the learning rate needs to be tuned for each task).  In the Appendix, Figure~\ref{fig:distinct_lr} shows that for both MPPI-CMA and CEM, ${(\alpha_{m},\alpha_{\Sigma})=(1,0.1)}$ outperforms a single shared rate $(1,1)$ or $(0.1,0.1)$ on both Go2Walk and HumanoidStandup. Interestingly,  we find that it is crucial to use distinct learning rates for the mean and covariance updates.

Third, the block-diagonal CMA update matches the performance of the full covariance matrix update (see Figure~\ref{fig:cov}). In other words, it appears that the lower computational cost of the block-diagonal CMA does not come at the price of lower performance. 

We refer the reader to the Appendix~\ref{appendix:TOresults} for more extensive results with all the other problems. Overall,
we find that CEM and MPPI-CMA perform the best. CEM tends to converge to better solutions. However, MPPI and MPPI-CMA tend to decrease the cost faster during the early iterations. This can be especially useful for MPC applications where the compute time is limited. In general, it appears that the key to the success of these algorithms is the use of CMA with different learning rates for mean and covariance.

\section{Policy optimization} \label{section:RL}

One of the main limitations of TO is that it is not always straightforward to use it to derive a controller. Indeed, performing MPC implies solving TO problems online, which can be extremely challenging due to limited compute time. Even though recent works were able to achieve impressive zero-order MPC demonstrations~\citep{li2024drop, xue2024full}, these performances have not yet matched those obtained by RL~\citep{handa2023dextreme, hoeller2024anymal}. The main strength of current RL approaches is that they rely solely on policy evaluation at runtime. All the computations are moved offline, where the goal is to search for a policy that maximizes performance across all possible initial conditions. This can be written:
\begin{align}
    F(\theta)  = \mathbb  E_{s} \left[ J(\theta, s) \right],  \label{eq:main_RL_cost}
\end{align}
where the expectation is taken over initial states $s$ and where $J(\theta, s)$ is the cost of the trajectory with initial state $s$ and with the controller $\pi_{\theta}$:
\begin{align}
    J(\theta, s) &=  \sum_{t=0}^\infty \gamma^t r(s_t, a_t),  \\
    \mbox{s.t. } \, s_0 = s, \,  & s_{t+1} = f_{\text{dyn}}(s_t, a_t) \, \mbox{ and } \, a_t = \pi_{\theta}(s_t). \nonumber
\end{align}
$\theta$ denotes the parameters of the policy (e.g. the weights of a neural network). $r$ is the reward, $a$ is the control action.
Note that we follow the RL formalism by trading the minimization of a cost function for the maximization of a reward. Furthermore, in contrast to TO, the optimization for each trajectory is done with an infinite horizon and a discount factor $\gamma$.
We do not consider stochastic dynamics for simplicity, as most robotic models are deterministic (e.g. locomotion or manipulation). However, the ideas presented in this section should extend to the stochastic dynamics case.
Because the policy is deterministic, RL algorithms aiming to solve formulation~\eqref{eq:main_RL_cost} are called Deterministic Policy Gradient (DPG) algorithms \citep{silver2014deterministic}.

\begin{remark}
The formulation~\eqref{eq:main_RL_cost} can easily be extended to encompass domain randomization by taking the expectation across multiple parameters of the environment. 
\end{remark}

Most RL algorithms rely on stochastic policies, therefore DPG algorithms are often introduced as a special case.  
While stochastic policies can be relevant in the context of games or partially observable settings, most robotic applications, such as locomotion or manipulation, can be solved with deterministic policies. In fact, in practice, many works that rely on RL algorithms derived with stochastic policies generally deploy a deterministic policy on the robot. Therefore, we chose to first introduce the deterministic case and then investigate how stochastic formulations can be understood as a random search aimed at solving the deterministic case.

There are two main sampling approaches to maximize~\eqref{eq:main_RL_cost}. The first one is to directly rely on the zero-order techniques, which implies sampling in the parameter space of a neural network, similarly to \cite{salimans2017evolution, mania2018simple}. The second approach, used in RL, is to leverage the structure of the problem to sample from the action space. 
In this work, we are interested in the latter one. 

\subsection{Deterministic Policy Gradient}

In the deterministic setting, the gradient can be estimated with the DPG theorem~\citep{silver2014deterministic}:
\begin{theorem}
The deterministic policy gradient theorem states that
\begin{align}\label{eq:deterministic_policy_gradient}
    \frac{\partial}{\partial \theta}  F(\theta) &=  \mathbb  E \left[ \sum_{t=0}^{\infty} \gamma^t \partial_{\theta} \pi_{\theta}(s_t) \partial_a Q(s_t, a_t) \right].
\end{align}
where the $Q$-function is defined as
\begin{align}
    &Q(s, a) =  \sum_{t=0}^\infty \gamma^t r(s_t, a_t) , \\
     \mbox{s.t. }  \, s_0 = s, \;& a_0 = a, \; s_{t+1} = f_{\text{dyn}}(s_t, a_t), \; a_t = \pi_{\theta}(s_t). \nonumber
\end{align}
\end{theorem}
Note that we follow the formalism of \cite{nota2019policy} and express the policy gradient as an expectation over entire trajectories rather than making the $\gamma$-discounted state distribution explicit.

One of the key challenges with the DPG is that the gradient can not be directly implemented in an algorithm because of the presence of the $Q$-function. Indeed, the $Q$-function cannot be computed analytically. One solution is to estimate it with function approximation at each gradient descent step. Algorithms relying on this mechanism are called actor-critics. The actor refers to the policy, while the critic refers to the $Q$-function. In practice, the $Q$-function is learned in a self-supervised way via the Bellman equation. There exist many techniques to estimate the critic efficiently between gradient updates of the actor; we refer the reader to \cite{suttonEdition2} for a comprehensive treatment of the subject. 

\begin{remark}
A key distinction between TO and policy optimization lies in the objective function: the policy optimization objective is averaged across various initial conditions. As a result, the objective cannot be computed analytically and must instead be estimated through sampling.  Consequently, gradient estimates are inherently noisy and introduce stochasticity into the optimization process~\citep{mandt2017stochastic}.
In light of our discussion on Langevin dynamics, this inherent stochasticity of RL algorithms could explain their empirical ability to avoid poor local minima.
\end{remark}

\subsection{Gaussian smoothing on the \texorpdfstring{$Q$}{Q}-function}

DDPG~\citep{lillicrap2015continuous} originally proposed to learn the $Q$-function with a neural network and directly differentiate it to obtain $\partial_a Q(s_t, a_t)$. 
This idea led to state-of-the-art RL algorithms such as TD3~\citep{fujimoto2018addressing}.
In light of the success of randomized smoothing, a natural idea is to use Gaussian smoothing to estimate the gradient of the Q-function and to use the following estimate for the gradient of the cost:
\begin{align}
\hspace{-0.25cm} \mathbb  E \left[\sum_{t=0}^{\infty} \gamma^t  \partial_{\theta} \pi_{\theta}(s_t)  (Q (s_t, a_t + \epsilon_t) - Q(s_t, a_t))\Sigma ^{-1}\epsilon_t  \right],\label{eq:custom_RS_on_Q}
\end{align}
where the expectation is taken with respect to both the initial condition and the Gaussian random variables~$\epsilon_t$.
This provides us with a simple variation of actor-critics based on the DPG theorem. Algorithm~\eqref{algo:rs-actor-critic} summarizes the procedure.

\begin{algorithm}[!ht]
\DontPrintSemicolon
\KwInput{$s, \theta, Q$}
    Compute action $a = \pi_{\theta}(s)$\;
    Sample i.i.d. random variables, $(\epsilon_k)_{k \in \{1,\dots,K\}}$\;
    ${\displaystyle \Delta\theta \gets \frac{1}{K} \sum_{k=1}^{K} (Q(s, a + \epsilon_k) - Q(s, a)) \partial_{\theta} \pi_{\theta}(s) \Sigma^{-1}  \epsilon_k} $\; \vspace{-4mm}
    Update the actor ${\displaystyle \theta \gets \theta + \alpha  \Delta\theta }$\;
    Update the critic, $Q$\;
\KwOutput{$\theta$}
\caption{Randomized smoothing - Actor-Critic update}
\label{algo:rs-actor-critic}
\end{algorithm}

We deliberately omit the rollout logic, as well as the interplay between the actor update and the critic update. This is because our focus is solely on modifying the actor’s gradient update in DPG algorithms. We do not aim to provide a novel actor-critic logic. As shown in the experimental section, this simple change can lead to competitive algorithms.
Note that in Algorithm~\ref{algo:rs-actor-critic}, the index $k$ in $\epsilon_k$ corresponds to the sample index, not the time index.

\begin{remark}
One way to understand this algorithm is to interpret the partial derivative of the policy $\partial_{\theta} \pi_{\theta}(s) $ as a linear mapping from the action space to the policy parameter space. 
Consequently, the algorithm can be understood as a way to map samples from the action space into samples from the policy parameter space. 
\end{remark}

An interesting variation of Algorithm~\ref{algo:rs-actor-critic} is to use the log-sum-exp transform and use exponential average weights in the actor update. More specifically, we can use the following update rule based on the log-sum-exp transform:
\begin{align}
\hspace{-0.25cm} \mathbb  E \left[\sum_{t=0}^{\infty} \gamma^t  \partial_{\theta} \pi_{\theta}(s_t) 
 \frac{   \exp\left(- \frac{1}{\lambda} Q (s_t, a_t + \epsilon_t) \right) \epsilon_t  )  }{ \mathbb   E \left[ \exp\left(- \frac{1}{\lambda} Q (s_t, a_t + \epsilon_t) \right) \right] } \right],\label{eq:custom_LSE_on_Q}
\end{align}
which leads to the following actor-critic update step.
\begin{algorithm}[!ht]
\DontPrintSemicolon
\KwInput{$s, \theta, Q$}
    Compute action $a = \pi_{\theta}(s)$\;
    Sample i.i.d. random variables, $(\epsilon_k)_{k \in \{1,\dots,K\}}$\;
    ${\displaystyle \Delta\theta \gets \    \frac{    \sum_{k=1}^{K} \exp\left(- \frac{1}{\lambda} Q (s, a + \epsilon_k) \right)  \epsilon_k   }{ \mathbb   \sum_{k=1}^{K} \exp\left(- \frac{1}{\lambda} Q (s, a + \epsilon_k) \right)  } } $\; 
    Update the actor ${\displaystyle \theta \gets \theta + \alpha  \Delta\theta }$\;
    Update the critic, $Q$\;
\KwOutput{$\theta$}
\caption{Log-Sum-Exp - Actor-Critic update}
\label{algo:lse-actor-critic}
\end{algorithm}

\subsection{Numerical experiments}

In this section, we investigate the benefits of estimating the gradient of the $Q$-function in DPG algorithms such as DDPG and TD3. Our implementation relies on CleanRL \citep{huang2022cleanrl} and only modifies the actor's update rule. Specifically, we investigate two smoothing approaches, the default randomized smoothing (RS) and the one relying on the log-sum-exp (LSE) transform. Consequently, we benchmark DDPG and TD3 against their smoothed counterparts: RS-DDPG, LSE-DDPG, RS-TD3, and LSE-TD3.
The performances are evaluated in seven MuJoCo environments.
For the sampling, we use 10 samples per update, with sampling noise following a Gaussian distribution with a standard deviation of $0.1$ (the covariance matrix is proportional to the identity matrix).
Our implementation is available online~\footnote{\url{https://github.com/ajordana/zoo-rob}}.

\begin{figure}[ht]
        \centering
        \includegraphics[width=0.5\textwidth]{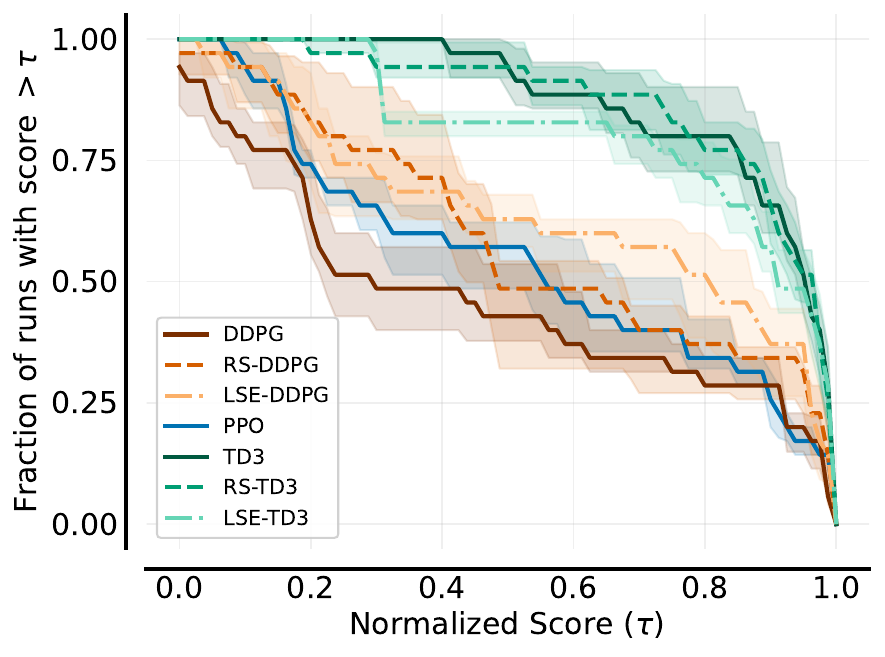}
        \caption{Fraction of runs with a normalized score above a given threshold ($\tau$). Higher is better. }
        \label{fig:compare_performance_profile_1}
\end{figure}

Five runs are performed for each test problem. Figure~\ref{fig:compare_performance_profile_1} presents the normalized score across all runs. The episodic return for each environment is provided in Appendix~\ref{appendix:RLresults}. 
The results indicate that both RS-DDPG and LSE-DDPG significantly outperform DDPG, which demonstrates the benefits of smoothing.  For TD3, the improvements are not as clear.
This could be explained by the fact that TD3 is already a very strong algorithm and that the margin for improvement on those benchmarks is limited. 
Additionally, our modifications were restricted to the actor's update; achieving state-of-the-art performance would most likely require extensive hyperparameter tuning of the actor-critic mechanism.
Nevertheless, the results show that deterministic policy gradient algorithms can benefit from smoothing.

In the following section, we will see how the novel RL algorithms proposed can help shed light on the connections between stochastic and deterministic RL policy gradient techniques typically used in robotics.

\subsection{Connection to the Reinforce algorithm}

One may ask how the update rule~\eqref{eq:custom_RS_on_Q} compares to the classical Reinforce algorithm~\citep{suttonEdition2}. First, let's recall how this algorithm is derived. In its most general form, Reinforce relies on the gradient of the following formulation:
\begin{align}~\label{eq:stochastic formulation}
   & J(\theta, s) =  \mathbb  E \left[ \sum_{t=0}^\infty \gamma^t  r_t(S_t, A_t) \right], \\ 
 \mbox{s.t. }\;\;  S_0 &= s \;,  A_t \sim  p_{\theta}( \; . \;  | S_t), \; S_{t+1} = f_{\text{dyn}}(S_t, A_t). \nonumber
\end{align}
We now use capitalized letters for actions and states to emphasize their stochastic nature. Indeed, the policy is no longer a deterministic function but a parameterized distribution over the space of possible actions. Note that the expectation is taken over random actions. 
\begin{theorem}    
Given $F(\theta) = \mathbb E_{s} [J(\theta, s_0)]$, the stochastic policy gradient theorem states that
\begin{align} 
    \frac{\partial}{\partial \theta} F (\theta) &=  \mathbb  E \left[ \sum_{t=0}^{\infty} \gamma^t Q(S_t, A_t) \frac{\partial}{\partial \theta} \log( p_{\theta}(A_t| S_t))  \right] .\nonumber 
\end{align}
where the expectation is taken over the initial condition and random actions.
\end{theorem}
Note that the definition of the $Q$-function is no longer the same as in the deterministic case. It is now defined as the expectation of stochastic rollouts starting at state $s$ with action~$a$. In practice, in most RL algorithms, the actions are sampled from a Gaussian probability distribution with mean $\pi_{\theta}(s)$ where $\pi_{\theta}$ is a neural network. While it is possible to also optimize the parameters of the covariance, we consider, for simplicity, a Gaussian distribution with a fixed covariance $\Sigma$. Hence, we have:
\begin{align}
 a \sim \mathcal{N}(\pi_{\theta}(s), \Sigma).
\end{align}
Therefore,
\begin{align}
 \frac{\partial}{\partial \theta} \log( p_{\theta}(a| s)) = \frac{\partial \pi_{\theta}(s)}{\partial \theta}^\top \Sigma ^{-1} (a - \pi_{\theta}(s)).
\end{align}
Consequently, we have: 
\begin{align}~\label{eq:reinforce_update}
   \frac{\partial}{\partial \theta} F (\theta)  = &  \mathbb  E \left[ \sum_{t=0}^{\infty} \gamma^t \left( Q(S_t, A_t) - V(S_t) \right) \right. \nonumber \\
& \left.  \frac{\partial \pi_{\theta}(S_t)}{\partial \theta}^\top \Sigma ^{-1} (A_t - \pi_{\theta}(S_t))\right].
\end{align}
The baseline $V(S_t)$ can be subtracted because the random variable $A_t - \pi_{\theta}(S_t)$ has a zero-mean conditioned on the state. Similarly to the randomized smoothing case, this induces invariance to translation by a constant of the reward function.
Interestingly, we recover a formula very similar to the policy gradient update with randomized smoothing on the Q-function (Equation~\eqref{eq:custom_RS_on_Q}). The only difference is that the rollouts are now stochastic. This changes two things. The first is that the $Q$-function and the partial derivatives are now evaluated at $S_t$ (instead of $s_t$ in Equation~\eqref{eq:custom_RS_on_Q}). The second is that the definition of the $Q$-function has changed. In Equation~\eqref{eq:custom_RS_on_Q}, the $Q$-function denotes deterministic rollouts. In contrast, it is now an expectation of the outcome of stochastic rollouts.
 
The additional stochasticity in Equation~\eqref{eq:reinforce_update} (as compared to Equation~\eqref{eq:custom_RS_on_Q}) can be understood as exploration noise.
One formal justification could be to understand the stochastic policy in Equation~\eqref{eq:stochastic formulation} as a smoothing operation in action space. Similarly to Gaussian smoothing, the smoothing allows one to derive the gradient of the cost with only $Q$-function evaluations. Consequently, the policy gradient theorem is analogous to the log-likelihood trick of Gaussian smoothing.

In the end, Reinforce can be seen as a way to solve the deterministic formulation introduced in Equation~\eqref{eq:main_RL_cost} without the simulator's gradients by sampling in the action space. Interestingly, the smoothing of the Reinforce formulation introduces additional stochasticity compared to Equation~\eqref{eq:custom_RS_on_Q}. It would be interesting to understand whether this extra stochasticity, which can be interpreted as exploration noise, is useful in practice. Intuitively, smoothing helps to simplify the problem, but adding too much noise might render the surrogate too different from the initial problem.
Future work could explore connections to other popular stochastic policy gradient algorithms, such as TRPO or PPO.

\section{Population based algorithms} \label{section:population}

In the past two sections, we have seen how commonly used methods in either trajectory optimization or RL rely on approximate gradient descent. In other words, these algorithms can be seen as instances of discretized stochastic Langevin gradient dynamic (Eq. \eqref{eq:slgd}). By design, the update schemes of these techniques are inherently local.

While theoretically, sampling-based algorithms based on Langevin dynamics can generate samples from any target probability distribution, it is well-known that, when applied to challenging multimodal distributions, the samples may fail to mix well and can exhibit strong autocorrelation~\citep{andrieu2003introduction}.
Consequently, it is worth studying algorithms that combine local search concepts with global ones, in the spirit of Algorithm~\ref{algo:blind_search}.

The most naive way to combine local and global search is to perform multiple local searches with diverse initial starting points. This is also known as Random Restart. Algorithm~\ref{algo:restart} summarizes the procedure.

\begin{algorithm}[!ht]
\DontPrintSemicolon
\KwInput{$x_0 \in \mathbb R^n$}
$x \gets x_0$\;
\While{stopping criterion is not met}{
Sample $\bar{x}$ in $\mathbb R^n$\;
$\tilde{x} \gets \text{Greedy Local Search}(\bar{x} )$\;
  \If{$f(\tilde{x}) < f(x)$ }{
    $x \gets \tilde{x} $\;
  }
}
\KwOutput{$x$}
\caption{Random Restarts}\label{algo:restart}
\end{algorithm}

A more sophisticated approach is to perform several local searches in parallel. 
One way to do this is to maintain a population of $N$ samples, generate a new population by applying local search, and then select the $N$ best samples. This procedure, called $(N + \lambda)$-ES, is one of the most popular Evolutionary Strategy (ES) algorithms. Algorithm~\ref{algo:es} summarizes this procedure.
Interestingly, there are two special cases:
(1 + 1)-ES is equivalent to Greedy Local Search (Algorithm~\ref{algo:local_search}) and
(1 + $\lambda$)-ES is equivalent to Predictive sampling~\citep{howell2022predictive} (Algorithm~\ref{algo:predictive_sampling}).

\begin{algorithm}[!ht]
\DontPrintSemicolon
\KwInput{Initial population of $N$ values: $x_1, \dots, x_N$}
\While{Stopping criterion is not met}{
    $D \gets \{x_1, \dots, x_N\}$\;
    \For{$i=1,\dots \lambda$}{
    $x \gets$ Sample uniformly from $\{x_1, \dots, x_N\}$\;
    $d \gets $ Sample a centered Gaussian\;
    Add $x+d$ to $D$\;
    }
$\{x_1, \dots, x_N\} \gets$ the $N$ elements in $D$ with smallest $f(x)$\;
}
\KwOutput{$x$}
\caption{$(N+ \lambda) $-ES}
\label{algo:es}
\end{algorithm}

A potential limitation of this type of approach is that many greedy local searches may converge to the same local minimum. 
Ideally, samples should coordinate with one another. An intuitive idea is to avoid having particles that are too close to one another.
One way to do so is to perform Stein Variational Gradient Descent (SVGD)~\citep{liu2016stein}. This formulation naturally encompasses a repulsion term to maintain a certain distance between particles. Although SVGD requires gradients, it is natural to use the concepts covered in the previous Sections to derive a gradient-free variation. For instance, \cite{braun2024stein} derived a gradient-free SVGD based on CMA-ES.
On the trajectory optimization side, \cite{lambert2020stein} applies SVGD to MPC.
On the RL side, \cite{liu2017stein} propose a policy gradient variation of SVGD.

\section{Discussion} \label{section:practical}

\subsection{Parallel computing} \label{section:parralel}

A common feature of zero-order TO and RL is that both require numerous function evaluations. To improve computational efficiency, a typical strategy is to parallelize these evaluations. In the machine learning community, parallel computing is well established with libraries such as JAX~\citep{jax2018github} and PyTorch~\citep{paszke2019pytorch}. 
However, in robotics, these libraries cannot be used directly, as each function evaluation relies on a simulator.
Indeed, the function evaluation in Equation~\eqref{eq:TO} or Equation~\eqref{eq:main_RL_cost} involves propagating the system dynamics forward in time. For this reason, parallel simulators such as MuJoCo XLA (MJX)~\citep{todorov2012mujoco}, MuJoCo Warp, and Isaac Sim~\citep{NVIDIA_Isaac_Sim} have been developed to support massively parallel rollouts on GPUs, which makes them especially suitable for sampling-based TO and RL.

On the algorithmic side, Evosax~\citep{evosax2022github} provides implementations of random search and popular ES algorithms based on JAX. These can then be used for zero-order MPC with Hydrax~\citep{kurtz2024hydrax}, which provides various examples of cost functions to control complex robots. Finally, CleanRL~\citep{huang2022cleanrl} offers simple and efficient implementations of widely used RL algorithms.

\subsection{Other practical considerations}

This work focuses on the mathematical foundations of the commonly used algorithms in TO and RL. However, there are many practical considerations to take into account in order to use these techniques to control robots. 

First, it is important to take into account that the modeling of robots is never perfectly accurate. For RL methods, domain randomization has become standard~\citep{tobin2017domain, peng2018sim}. This simple idea consists of randomizing key uncertain model parameters (e.g., friction coefficient, actuation parameters, latencies). Similarly, one can directly model sensor noise by randomizing the sensors output. This is often a crucial step to bridge the sim-to-real gap. For sampling-based MPC, it is also possible to perform online domain randomization by directly randomizing the model parameters while rolling out the model to compute the cost function in Equation~\eqref{eq:TO}~\citep{abraham2020model, kurtz2024hydrax}.

Additionally, in the case of sampling based MPC, it is crucial to consider the computation time. While the compute-time bottleneck in MPC for robotics is a well-known problem~\citep{wensing2023optimization}, it is even more relevant in the context of sampling-based optimization for the simple reason that zero-order optimization typically requires much more iterations than traditional methods based on gradient information. Indeed, while gradient-free techniques are very attractive as they naturally handle the discontinuities inherent to contact, it is important to remember that this comes at the cost of higher compute time compared to traditional gradient-based techniques. To mitigate this problem, the use of splines is often necessary to reduce the dimensionality of the problem. Moreover, in practice, to further reduce compute time, practitioners typically operate with very short horizon lengths~\citep{xue2024full, crestaz2025td}, which can be problematic for the stability of the system~\citep{mayne2014model}. 
Furthermore, MPC should ideally re-plan at each control step using a receding horizon; however, the high compute times are often longer than the control step time. This forces practitioners to execute multiple steps from an optimized trajectory, which can be problematic for guaranteeing stability~\citep{mayne2014model}. 
Lastly, another limitation of gradient-free approaches is that they typically struggle to converge to high precision, which can be an issue in practice, as low accuracy solutions in MPC loops can lead to vibrations or failures. In practice, practitioners have to carefully tune the trade-off between solution quality (number of iterations), horizon length, and re-planning frequency. Intuitively, the horizon length and number of iterations should be as high as possible, provided that the compute time allows for a reasonable re-planning rate (50Hz or more for dynamic tasks).
A potential way to circumvent these challenges is to use TO offline to generate data to train a policy~\citep{liu2025opt2skill, dhedin2026dynaretarget}.

One of the big challenges of sampling-based MPC is the inclusion of constraints. In practice, hard constraints are an essential requirement for designing safe systems.
Furthermore, beyond safety, the ability to incorporate constraints is important in terms of cost design. Without constraints, practitioners have to describe complex behaviors by adding terms to the cost (or reward), which can lead to challenging cost tuning (or reward engineering). The ability to describe the task with a mix of cost (or reward) terms and constraint terms can be helpful in practice~\citep{chane2024cat}. 
While constrained MPC is a well-established practice, the incorporation of constraints in real-time with gradient-free optimization appears to be more challenging. 
While there exist gradient-free optimization methods that handle nonlinear constraints~\citep{conn2009introduction}, it is unclear whether they can be made real-time on complex robots with meaningful horizon lengths. Indeed, we are not aware of any successful demonstrations of constrained sampling-based MPC on complex robots. Overall, providing formal guarantees with sampling-based approaches remains an open problem.

\section{Conclusion}
In this work, we have demonstrated how random search provides a unifying perspective on zero-order algorithms commonly used in robotics. Our empirical study shows that the CMA update improves performance and simplifies the tuning of the learning rate. In practice, we find that the CEM and MPPI-CMA algorithms perform the best, and that block-diagonal covariance updates can match the performance of the full covariance update (at a lower algorithmic complexity). We also discussed theoretical concepts that help explain why sampling-based zero-order techniques can escape local minima.
Leveraging this understanding, we proposed novel and competitive RL algorithms. These algorithms are only examples of the potential outcomes enabled by this unified viewpoint. Table~\ref{tab:Summary} summarizes the main algorithm covered in this work.
In the future, we hope this tutorial will help the robotics community tackle open challenges such as constrained zero-order optimization and the search for global solutions. 

\begin{table}[ht!]
    \centering
    \begin{tabular}{|c|c|}
    \hline
     Algorithm & Analytical update     \\
    \hline
    MPPI (Alg.~\ref{algo:MPPI}) & Eq.~\eqref{eq:update_analytical_MPPI}  \\
    \hline
MPPI-CMA (Alg.~\ref{algo:CMA}) &    Eq.~\eqref{eq:MPPI-CMA_in_expectation}  \\
    \hline
    DDPG & Eq.~\eqref{eq:deterministic_policy_gradient}  \\
    \hline
    RS-DDPG (Alg.~\ref{algo:rs-actor-critic}) & Eq.~\eqref{eq:custom_RS_on_Q}  \\
    \hline
    LSE-DDPG (Alg.~\ref{algo:lse-actor-critic}) & Eq.~\eqref{eq:custom_LSE_on_Q}  \\
    \hline
    \end{tabular}
    \caption{Summary of all the algorithm covered as well as their corresponding analytical update (i.e. the update before performing Monte Carlo approximation)}
    \label{tab:Summary}
\end{table}

\begin{acks}
We sincerely thank Frederike Dümbgen and Olivier Sigaud for their constructive feedback, which greatly improved this paper. This work was supported in part by the US National Science Foundation under Grants 2222815 and 2315396, by gifts from Wandercraft, Ultra, NVIDIA, Infinite Machines and in part by the French AI Interdisciplinary Institute ANITI. ANITI is funded by the France 2030 program under Grant Agreement number ANR-23-IACL-0002.
\end{acks}

\bibliographystyle{SageH}
\bibliography{biblio}

\appendix


\section{Connection between the greedy local search and the Metropolis-Hastings algorithm}\label{appendix:GLSandMALA}

\cite{toussaint2024nlp} shows how the Metropolis adjusted Langevin algorithm can be interpreted as a line search with the Armijo condition. Similarly, in the trivial case where the step size $\alpha_k=0$ in Equation~\eqref{eq:slgd}, we can recover a connection to Algorithm~\ref{algo:local_search}. The idea of the Metropolis-Hastings algorithm~\citep{andrieu2003introduction} is to accept a step from $x$ to $x'$ with probability:
\begin{align}
    \min \left( 1, \frac{p(x')}{p(x)}\frac{q(x|x')}{q(x'|x)}\right).
\end{align}
If $\alpha_k=0$ in Equation~\eqref{eq:slgd}, then,
\begin{align}
    q(\cdot|x) \propto \mathcal{N} (x, \gamma^2 I).
\end{align}
Given Equation~\eqref{eq:exp_energy_eq}, the acceptance probability of the Metropolis-Hastings algorithm~\citep{andrieu2003introduction} is 
\begin{align}
     \min \left( 1, \frac{p(x')}{p(x)}\frac{q(x|x')}{q(x'|x)}\right)
 =  \min \left( 1, e^{-\left(f(x') - f(x)\right)} \right) .\nonumber
\end{align}
Interestingly, $e^{-\left(f(x') - f(x)\right)} > 1$ implies $f(x') < f(x)$. Hence, one could easily transform the Greedy Local Search (Algorithm~\ref{algo:local_search}) into an MCMC method with the Metropolis-Hastings algorithm~\citep{andrieu2003introduction}. Indeed, in Algorithm~\ref{algo:local_search}, we could accept the step $x' = x+d$ with a probability of one in the case of a decrease and allow for an increase with a probability of $e^{-\left(f(x') - f(x)\right)}$.

\section{About the natural gradient}\label{appendix:natural_gradient}

\subsection{Deriving the natural gradient}

The idea of the natural gradient~\citep{amari1998natural} is to find the steepest descent according to an arbitrary metric $d$. 
More precisely, the natural gradient looks for a small step $h$ satisfying:
\begin{align}
\min_{h} & f(x) + \nabla f(x)^T h, \label{eq:natural_gradient}\\
\mbox{s.t. } & d(x + h, x) = \epsilon^2. \nonumber
\end{align}
where $\epsilon$ is a small scalar. Typically, $d$ is such that (at least locally):
\begin{align}
    d(x_1, x_2) = \frac{1}{2} (x_2 - x_1)^T A  (x_2 - x_1).
\end{align}
Let's write the Lagrangian of the problem~\eqref{eq:natural_gradient}
\begin{align}
   \mathcal{L}(h, \lambda) = f(x) + \nabla f(x)^T h - \lambda (\frac{1}{2}h^T A h - \epsilon ^ 2) .
\end{align}
We have:
\begin{align}
    \nabla_h \mathcal{L}(h, \lambda) = 0 &\iff  \nabla f(x)  - \lambda A h = 0 \nonumber \\
    & \iff h = \frac{1}{\lambda} A^{-1} \nabla f(x).
\end{align}
The sign of $\lambda$ can be recovered with the second order optimality condition~\citep{nocedal1999numerical}:
\begin{align}
    \nabla^2_{hh} \mathcal{L}(h, \lambda) \succeq 0 \implies - \lambda A  \succeq 0 \implies \lambda \leq 0.
\end{align}
Hence, the natural gradient step is of the form:
\begin{align}
    h \propto - A^{-1} \nabla f(x). \label{eq:nat_grad_formula}
\end{align}

\subsection{Recovering the Fisher matrix}

In the context where $\theta$ characterizes a probability distribution, a typical choice of distance is:
\begin{align}
 \hspace{-0.2cm}   \operatorname{KL}(p_{\theta + h} || p_{\theta}) = \int_x p_{\theta + h}(x) \log\left(\frac{ p_{\theta + h}(x)}{ p_{\theta }(x)}\right) dx . \label{eq:kl} 
\end{align}
Locally, we can show that this metric is quadratic. Let's look at the partial derivatives of Equation~\eqref{eq:kl} with respect to $h$:
\begin{align}
    &\frac{\partial \operatorname{KL}(p_{\theta + h} || p_{\theta})}{\partial h} =\int_x \partial_h p_{\theta + h}(x) \log\left(\frac{ p_{\theta + h}(x)}{ p_{\theta }(x)}\right) dx . \nonumber
\end{align}
As $\int_x   p_{\theta + h}(x) dx = 1$ implies $\partial_h  \int_x   p_{\theta + h}(x) dx = 0$ for all $h$. Hence,
\begin{align}
    \frac{\partial \operatorname{KL}(p_{\theta + h} || p_{\theta})}{\partial h} (h=0) =  0.
\end{align}
Then, we can compute the Hessian with respect to $h$:
\begin{align}
 & \hspace{-0.3cm}  \frac{\partial^2 \operatorname{KL}(p_{\theta + h} || p_{\theta})}{\partial^2 h} = \int_x \partial^2_h p_{\theta + h}(x) \log\left(\frac{ p_{\theta + h}(x)}{ p_{\theta }(x)}\right) dx \nonumber\\
    & \;\;\;\;\;\;\;\; +  \int_x \partial_h p_{\theta + h}(x)^T \partial_h p_{\theta + h}(x)  \frac{1}{ p_{\theta +h }(x)} dx.
\end{align}
And we find that:
\begin{align}
   & \frac{\partial^2 \operatorname{KL}(p_{\theta + h} || p_{\theta})}{\partial^2 h} (h=0) \\
    &= \int_x \partial_h p_{\theta }(x)^T \partial_h p_{\theta }(x)  \frac{1}{ p_{\theta}(x)} dx \nonumber \\
    &= \int_x \partial_{\theta} \log(p_{\theta}(x))^T \partial_{\theta} \log(p_{\theta}(x)) p_{\theta}(x) dx \nonumber \\
    &=  \mathbb E_{X \sim p_{\theta}} \left[  \partial_{\theta} \log(p_{\theta}(X))^T \partial_{\theta} \log(p_{\theta}(X)) \right] \nonumber = F(\theta),
\end{align}
where $F(\theta)$ is the Fisher matrix. In the end, 
\begin{align}
     \operatorname{KL}(p_{\theta + h} || p_{\theta}) = \frac{1}{2} h^T F(\theta) h + o(\vert\vert h \vert\vert^2).
\end{align}

Let's consider the specific case where $\theta = m$ and:
\begin{align} \nonumber
    p_{\theta}(x) = \frac{1}{\kappa} \exp\left(- \frac{1}{2}(x-m)^T \Sigma^{-1} (x-m)\right).
\end{align}
We have
\begin{align}
    \partial_{\theta} \log(p_{\theta}(x)) = (x-\theta)^T \Sigma^{-1} .
\end{align}
Therefore, the Fisher matrix reads:
\begin{align}
    F(\theta) &= \mathbb E \left[ \Sigma^{-1} (x-\theta) (x-\theta)^T \Sigma^{-1}  \right]\\
      &=  \Sigma^{-1} \mathbb  E \left[ (x-\theta) (x-\theta)^T \ \right]  \Sigma^{-1} \nonumber \\
      &=  \Sigma^{-1} . \nonumber
\end{align}
Hence, in that case, the Fisher matrix is equal to the inverse of the covariance matrix.
The case where $\theta = (m, \Sigma)$ can be found in \cite{akimoto2010bidirectional}.

\subsection{An alternative view}

\cite{wagener2019online} derives MPPI as an instance of mirror descent, a concept from online learning~\citep{beck2003mirror}. More precisely, given a strongly convex function $\psi$, \cite{wagener2019online} defines the step $h$ with the following formula:
\begin{align}
    h = \arg\min_h   \nabla f(x)^T (x + h ) + \frac{1}{\eta} D_\psi(x+h, x)  
\end{align}
with $\eta>0$ and where $D_\psi(x, y)$ is the Bregman-like distance defined as:
\begin{align}
    D_\psi(x, y) = \psi(x) - \psi(y) - (x-y)^T \nabla \psi(y)
\end{align}
If we choose, $\psi(x) = x^T A x$, we have:
\begin{align}
    D_\psi(x, y)  &= (x-y)^T A (x- y)
\end{align}
For this distance, the step reads:
\begin{align}
    h &= \arg\min_h   \nabla f(x)^T (x + h ) + \frac{1}{\eta} h^T A h \nonumber\\
    &= \eta A^{-1} \nabla f(x)
\end{align}
Hence, the obtained step is equivalent to Eq~\eqref{eq:nat_grad_formula}. By setting~$A$ to be the Fisher matrix, \cite{wagener2019online} recovers MPPI from online learning tools.

\section{About the connection between the log-sum-exp and risk-sensitive control}\label{appendix:risk}

\begin{figure}[ht!]
    \centering
    \includegraphics[width=\linewidth]{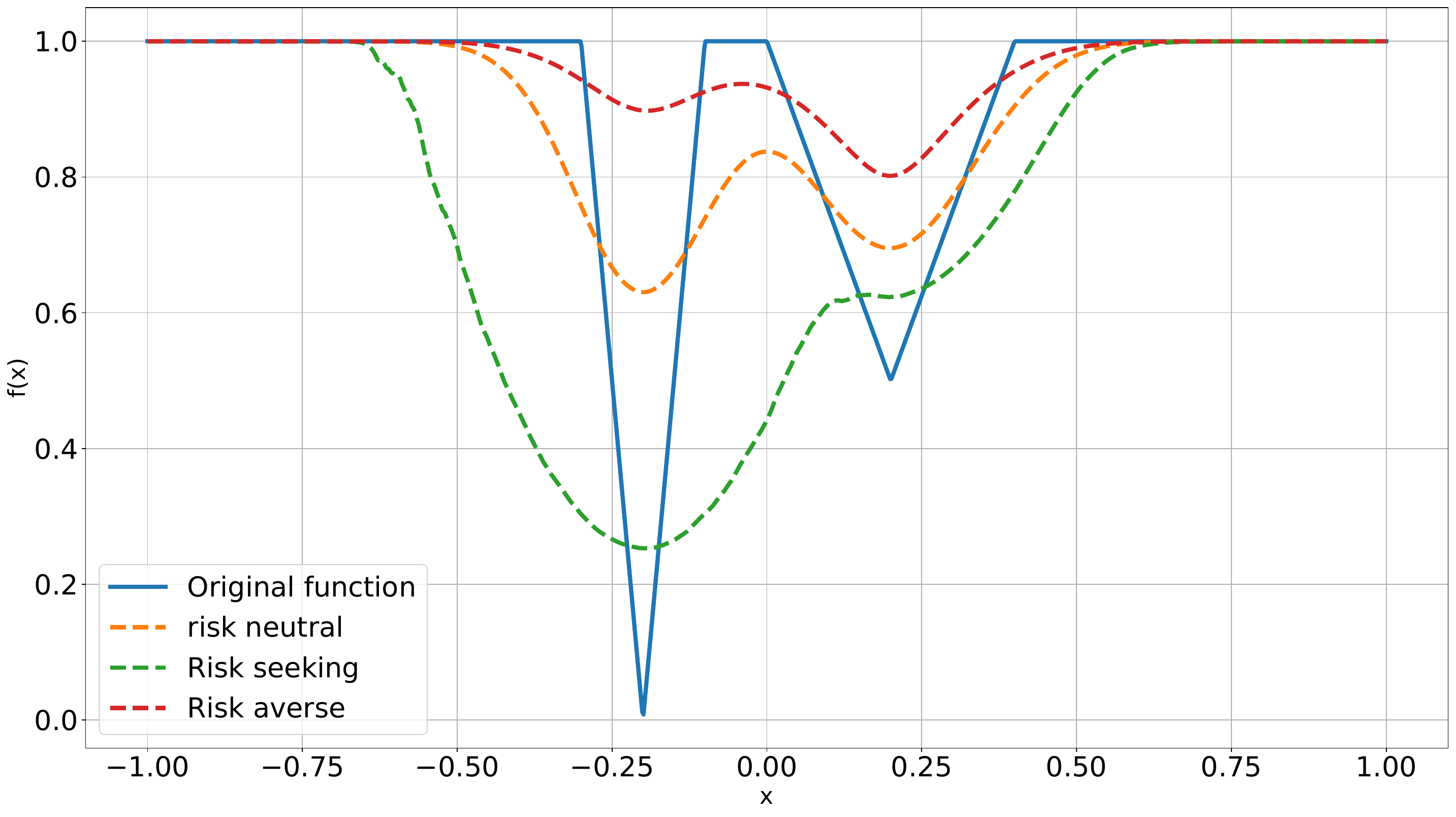}
    \caption{Risk-seeking and Risk-averse transformation}
    \label{fig:risk-illustration}
\end{figure}

Interestingly, the surrogate defined in Equation~\eqref{eq:mppi_surrogate} resembles the risk-seeking control formulation~\citep{whittle1981risk}. At first, this may seem counter-intuitive, as in practice, we prefer designing risk-averse controllers rather than risk-seeking ones. However, from an optimization perspective, the risk-seeking formulation has the advantage of giving more importance to lower values. Figure~\ref{fig:risk-illustration} illustrates the landscape for each formulation. The risk-neutral formulation corresponds to the default randomized smoothing. One way to interpret these landscapes is to see the smoothing noise as a disturbance from the environment. On one hand, the risk-seeking formulation expects the noise to act in its favor by shifting the state towards the global solution.  Therefore, it lowers the function value in the vicinity of the narrow global minima. On the other hand, the risk-averse formulation expects the noise to be adversarial. Therefore, it increases the function value (as compared to the neutral formulation) around the narrow global minima.

\section{Block diagonal CMA}\label{appendix:block}

We provide a proof for theorem~\ref{th:block-diagonal}. 

\begin{proof}
We know that:
\begin{align*}
- \lambda^{-1}   F(\theta)^{-1}\nabla J_{\lambda}(\theta) = \mathbb E \left[  w_f(z) F(\theta)^{-1} \nabla_{\theta} \log (p_{\theta}(z))  \right]
\end{align*}

Let's now derive $F(\theta)^{-1} \nabla_{\theta} \log (p_{\theta}(z))$
\begin{align}    
\log p_{\theta}(z)=& - \frac{d}{2} \log(2\pi) -\frac{1}{2} \log \operatorname{det}(\Sigma) \nonumber \\
& - \frac{1}{2} (z-m)^T \Sigma^{-1} (z-m) \nonumber \\
 =&  - \frac{d}{2} \log(2\pi) - \frac{1}{2} \sum_{t=1}^T  \log \operatorname{det}(\Sigma_t) \nonumber \\ &- \frac{1}{2} \sum_{t=1}^T  (z_t-x_t)^T \Sigma_t^{-1} (z_t-x_t) 
\end{align}

We have:
\begin{align}
  \nabla_{x_t}  \log p_{\theta}(z) = \Sigma_t^{-1} (z_t-x_t) 
\end{align}
and as $\frac{\partial  }{\partial A} z A ^{-1} z
=
-
A^{-1}zz^\top A^{-1}$, 
\begin{align}
  \frac{\partial}{\partial \Sigma_t}  \log p_{\theta}(z) = \frac{1}{2} \left( \Sigma_t^{-1} (z_t - x_t) ( z_t-x_t ) ^T  \Sigma_t^{-1} - \Sigma_t^{-1} \right)  \nonumber 
\end{align}
With $\theta$ defined as in \ref{eq:theta_for_block}, $\nabla_{\theta} \log p_{\theta}(z)$ reads:
\begin{align}
     & \begin{bmatrix}\Sigma_1^{-1} (z_1-x_1) \\
    \vdots\\
    \Sigma_T^{-1} (z_T-x_T) \\
    \frac{1}{2}   \operatorname{vec} \left( \Sigma_1^{-1} (z_1 - x_1) ( z_1-x_1 ) ^T  \Sigma_1^{-1} - \Sigma_1^{-1} \right) \\
    \vdots \\    
 \frac{1}{2}   \operatorname{vec} \left( \Sigma_T^{-1} (z_T - x_T) ( z_T - x_T) ^T  \Sigma_T^{-1} - \Sigma_T^{-1} \right) 
\end{bmatrix} \nonumber
\end{align}

Now, let's derive the Fisher Information Matrix (FIM). For a Gaussian distribution of the form $\mathcal{N}(m(\theta), \Sigma(\theta))$, the following identity holds:
\[
F_{i,j} = \frac{\partial m}{\partial \theta_i}^T \Sigma^{-1} \frac{\partial m}{\partial \theta_j} + \frac{1}{2} \operatorname{tr}(\Sigma^{-1} \frac{\partial \Sigma}{\partial \theta_i} \Sigma^{-1} \frac{\partial \Sigma}{\partial \theta_j})
\]
In our case, the first $Td$ elements of $\theta$ correspond to the elements of the mean, while the remaining elements correspond to the elements of the blocks of the covariance matrix.
There are three cases to study:
\begin{itemize}
    \item $1 \leq i, j \leq Td$, then $\theta_i$ corresponds to some $({m_p})_q$, and $\theta_j$ corresponds to $({m_{q'}})_{q'}$ (the first index corresponds to time and the second to the coordinate of the vector). This implies:
    \[
F_{i,j} =  [\Sigma^{-1}]_{i, j} 
\]
    \item $Td < i, j $, then there exist $(t_i, k,l)$ and $(t_j, m,n)$ such that $\theta_i = (\Sigma_{t_i})_{k,l}$ and $\theta_j = (\Sigma_{t_j})_{m,n}$.
Let's denote $E$ the indicator matrices of size $(d,d)$ and $\bar E$ the indicator matrices of size $(Td,Td)$. We have
\begin{align}
   & F_{i,j} =  \frac{1}{2} \operatorname{tr}(\Sigma^{-1} \bar E_{dt_i + k,dt_i+l} \Sigma^{-1} \bar E_{dt_j + m,dt_J+n})  \nonumber \\
    &= \frac{1}{2} \sum_{t=1}^T \operatorname{tr}(\Sigma_t^{-1} E_{k,l}\delta_{t=t_i}  \Sigma_t^{-1}E_{m,n}\delta_{t=t_j})  \nonumber\\
    &= \frac{1}{2} \sum_{t=1}^T \operatorname{tr}(\Sigma_t^{-1} E_{k,l} \Sigma_t^{-1}E_{m,n}) \delta_{t=t_i} \delta_{t=t_j}  \nonumber 
\end{align}
\begin{itemize}
    \item Case 1: $t_i \neq t_j$:
\end{itemize}
\begin{align}
F_{i,j} &=  0
\end{align}
\begin{itemize}
    \item Case 2: $t_i = t_j$:
\end{itemize}
\begin{align}
F_{i,j} &=  \frac{1}{2} \operatorname{tr}(\Sigma_{t_i}^{-1} E_{k,l} \Sigma_{t_i}^{-1} E_{m,n}) 
\end{align}
 As $\operatorname{tr}(D^TE_1^TDE_2) = \operatorname{vec}(E_1)^T (D \otimes D) \operatorname{vec}(E_2)$ \citep{harville1998matrix} and as $\Sigma_{t_i}^{-1}$ is symmetric, we have:
 \begin{align} 
F_{i,j} &= \frac{1}{2} \operatorname{vec}(E_{k,l})^T (\Sigma_{t_i}^{-1} \otimes \Sigma_{t_i}^{-1}) \operatorname{vec}(E_{m,n})
 \end{align}
    \item $1 \leq i \leq Td$ and $Td <j $, then $\theta_i = (m_p)_q $ and there exists a triplet $(t, m,n)$ such that $\theta_j = (\Sigma_t)_{m,n}$. This implies:
    \[F_{i, j} = 0\]
\end{itemize}
Hence, we have $ F(\theta) $ is equal to
\begin{align*}
\begin{bmatrix}
  \Sigma_1^{-1}   &&&&& \\
    &\ddots&&&& \\
    &&    \Sigma_T^{-1} &&& \\[2mm]
    &&& \frac{1}{2}\Sigma_1^{-1} \otimes \Sigma_1^{-1} && \\
    &&&& \ddots&\\
    &&&&& \frac{1}{2}\Sigma_T^{-1} \otimes \Sigma_T^{-1}
    \end{bmatrix}
\end{align*}
and as $(A \otimes A)^{-1} = (A^{-1} \otimes A^{-1})$, we have:

\begin{align*}
    F(\theta)^{-1} &= \begin{bmatrix}
  \Sigma_1   &&&&& \\
    &\ddots&&&& \\
    &&    \Sigma_T &&& \\[2mm]
    &&& 2\Sigma_1 \otimes \Sigma_1 && \\
    &&&& \ddots&\\
    &&&&& 2\Sigma_T \otimes \Sigma_T 
    \end{bmatrix}
\end{align*}

And as $\operatorname{vec}(ABA^T) = (A \otimes A) \operatorname{vec}(B)$ \citep{harville1998matrix}, $F(\theta)^{-1} \nabla_{\theta} \log p_{\theta}(z)$, is equal to:
\begin{align*}
 \begin{bmatrix}z_1-x_1 \\
    \vdots\\
    z_T-x_T \\
 \operatorname{vec} \left(  (z_1 - x_1) ( z_1-x_1 ) ^T  - \Sigma_1 \right) \\
    \vdots \\    
 \operatorname{vec} \left( (z_T - x_T) ( z_T - x_T) ^T  - \Sigma_T \right) 
\end{bmatrix}
\end{align*}
This concludes the proof.    
\end{proof}

\section{The Cross Entropy Method (CEM)} \label{appendix:CEM}

Algorithm~\ref{algo:CEM} summarizes the main steps of CEM.

\begin{algorithm}[h!]
\DontPrintSemicolon
\KwInput{$x_0 \in \mathcal{X}, \Sigma_0, K_e$}
$ (x, \Sigma) \gets (x_0, \Sigma_0)$\;
\While{stopping criterion is not met}{
    Sample $x_{1:K} \sim \mathcal{N}(x, \Sigma)$ \;
    Sort $x_{1:K}$ such that $f(x_1) \leq f(x_2) \leq \dots \leq f(x_K)$       \;
    $x \gets (1-\alpha_{m} )x + \alpha_{m} \sum\limits_{k=1}^{K_e} \frac{1}{K_e} x_k$  \;
    $\Sigma \gets (1-\alpha_{\Sigma} )\Sigma + \alpha_{\Sigma}  \sum\limits_{k=1}^{K_e} \frac{1}{K_e} (x_k - x)(x_k - x)^\top$ 
}
\KwOutput{$x$}
\caption{Cross-Entropy Method (CEM)}
\label{algo:CEM}
\end{algorithm}

\section{Additional TO results}~\label{appendix:TOresults}

\begin{figure*}
\centering
\begin{subfigure}{0.5\textwidth}\centering
  \includegraphics[width=\linewidth]{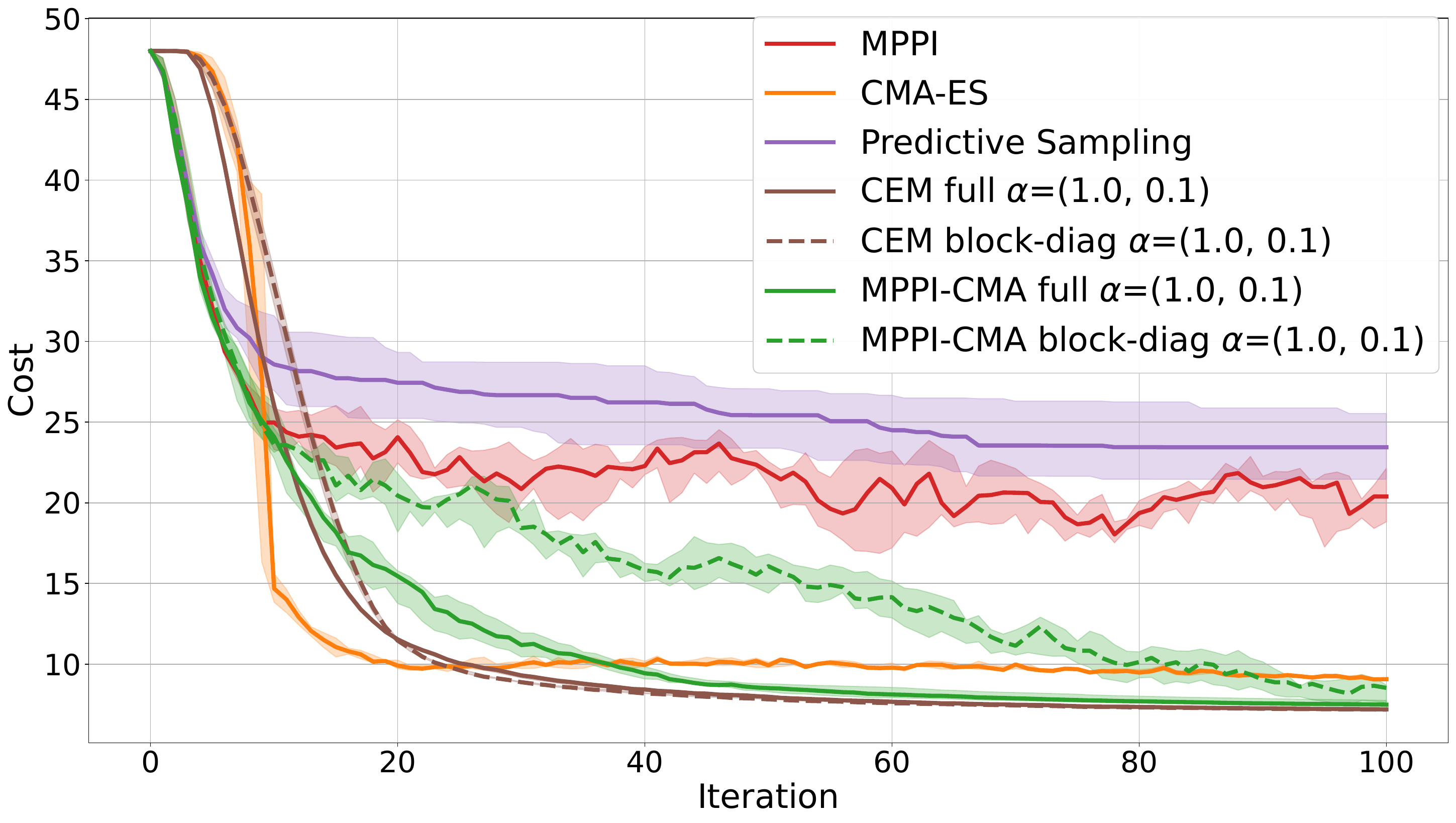}
  \caption{CartPole}\end{subfigure}\hfill
\begin{subfigure}{0.5\textwidth}\centering
  \includegraphics[width=\linewidth]{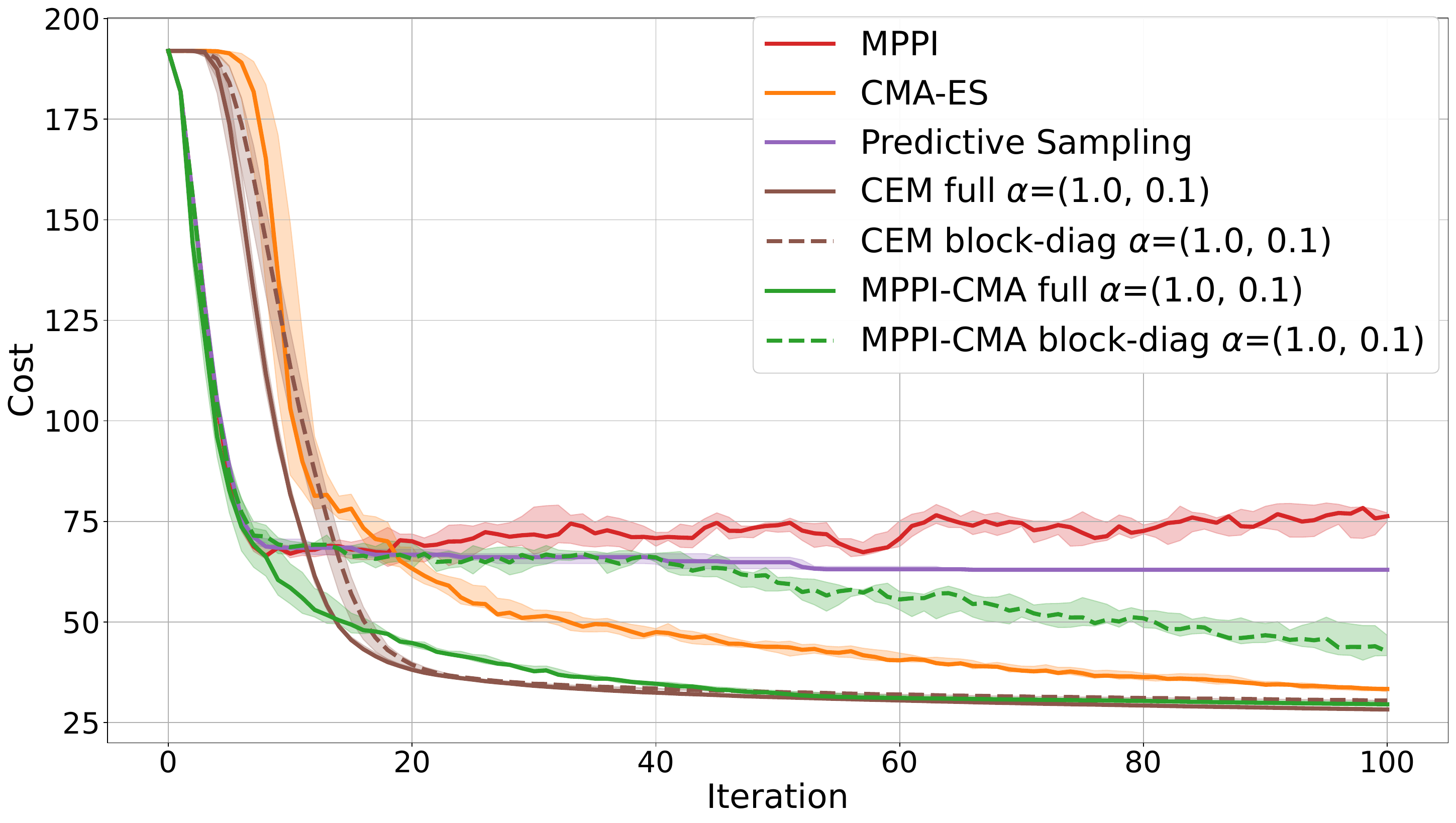}
  \caption{DoubleCartPole}\end{subfigure}\\[1.2ex]
\begin{subfigure}{0.5\textwidth}\centering
  \includegraphics[width=\linewidth]{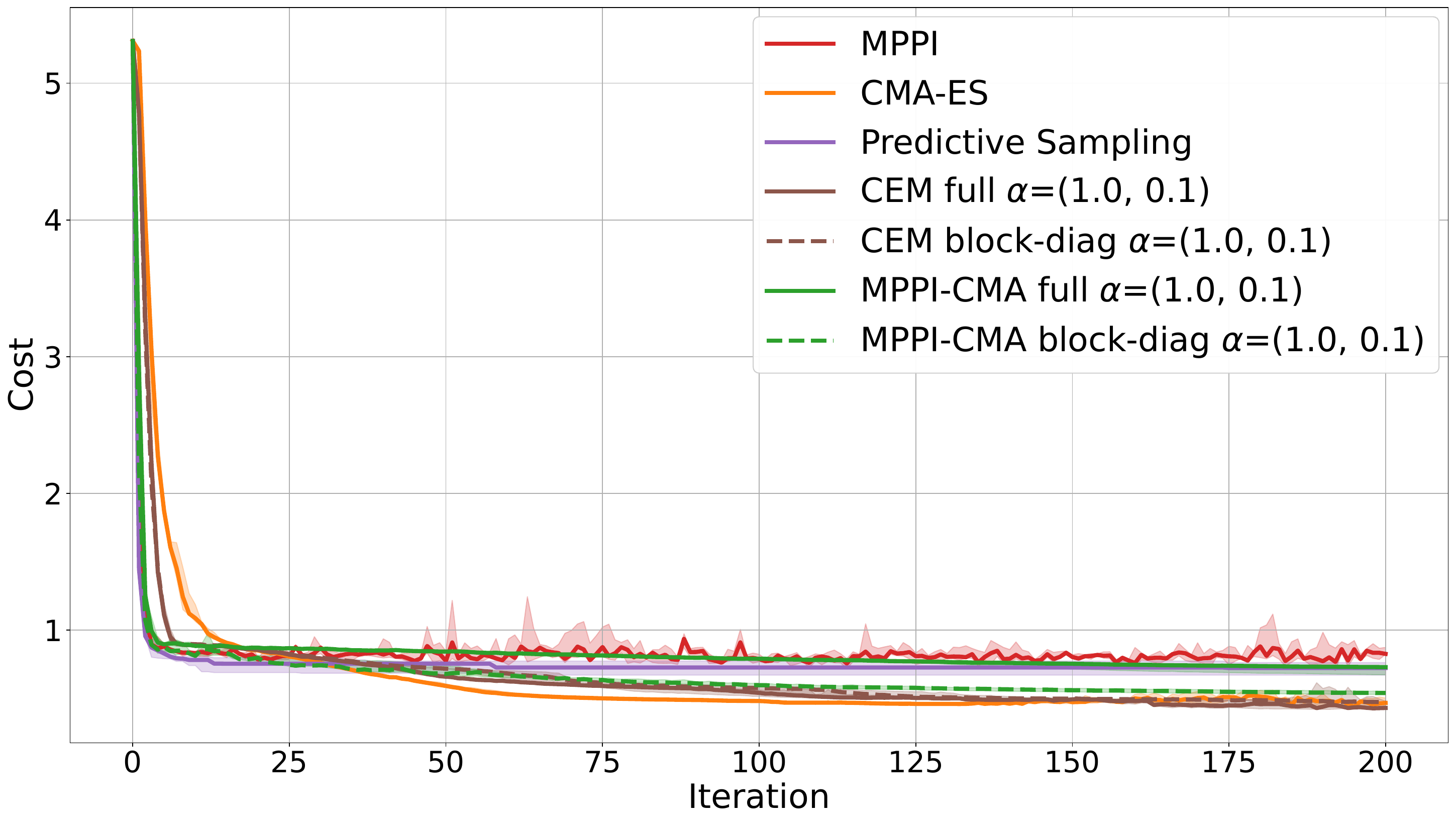}
  \caption{PushT}\end{subfigure}\hfill
\begin{subfigure}{0.5\textwidth}\centering
  \includegraphics[width=\linewidth]{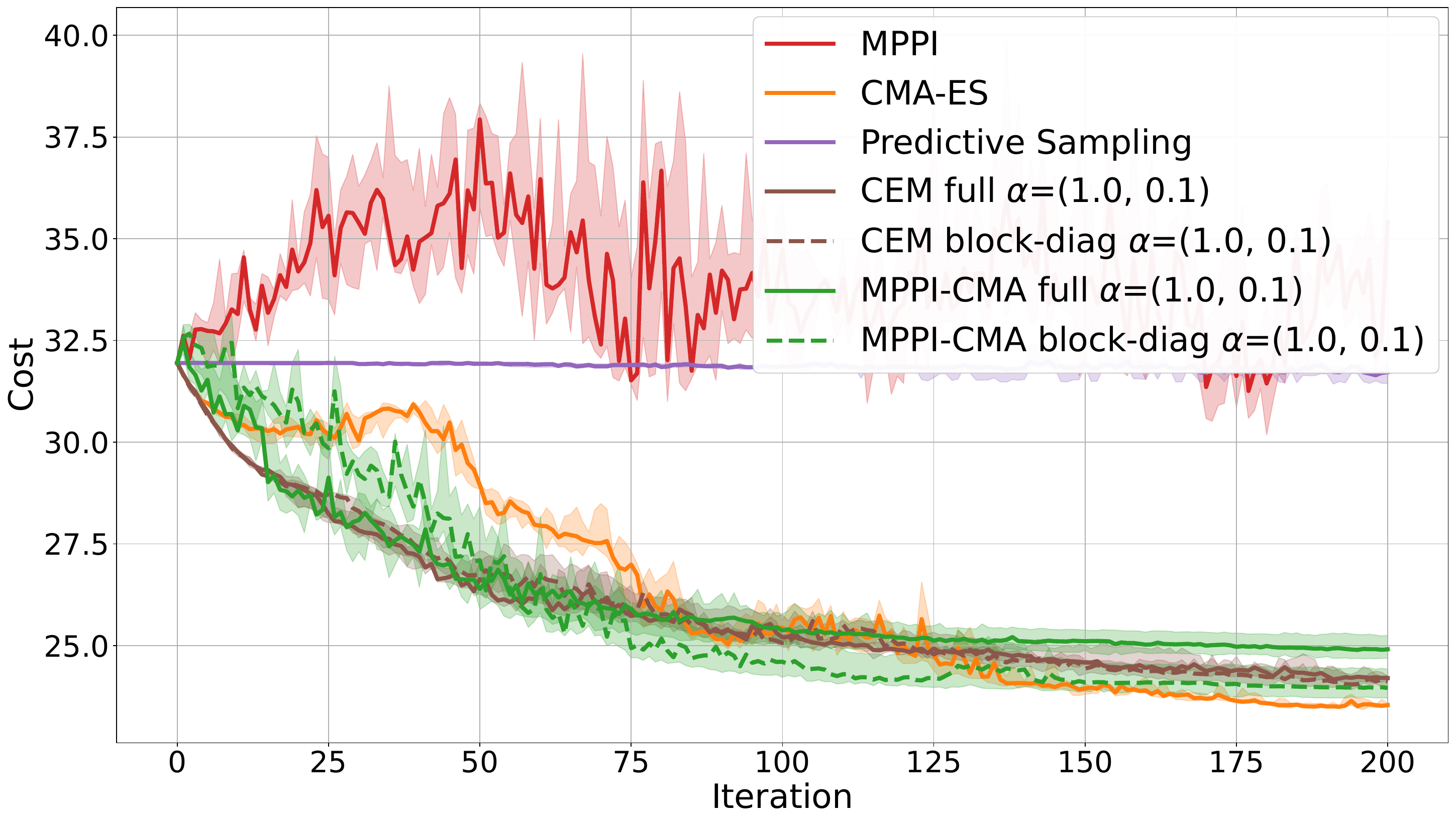}
  \caption{HumanoidStandup}\end{subfigure}\\[1.2ex]
\begin{subfigure}{0.5\textwidth}\centering
  \includegraphics[width=\linewidth]{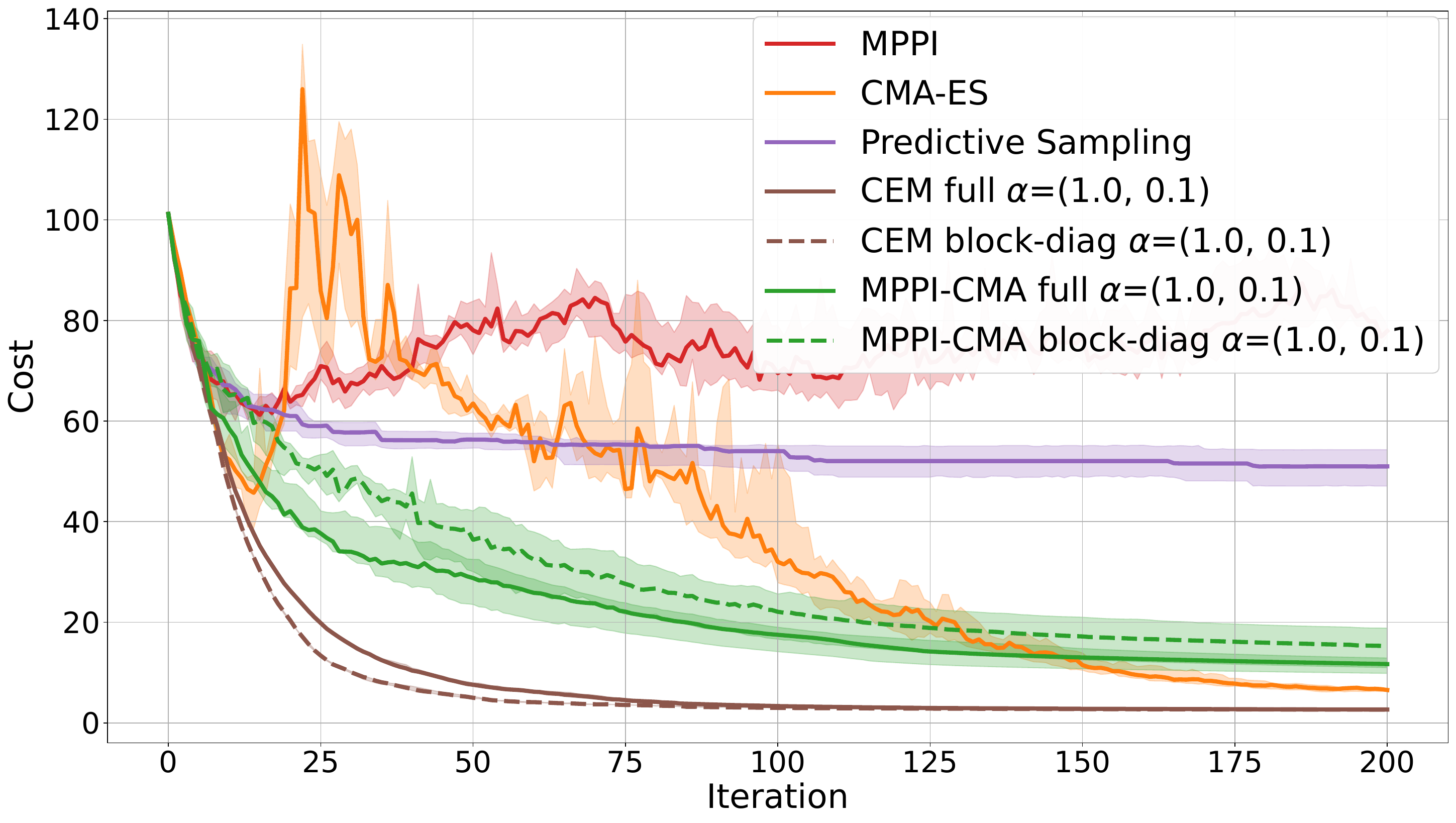}
  \caption{HumanoidBalance}\end{subfigure}\hfill
\begin{subfigure}{0.5\textwidth}\centering
  \includegraphics[width=\linewidth]{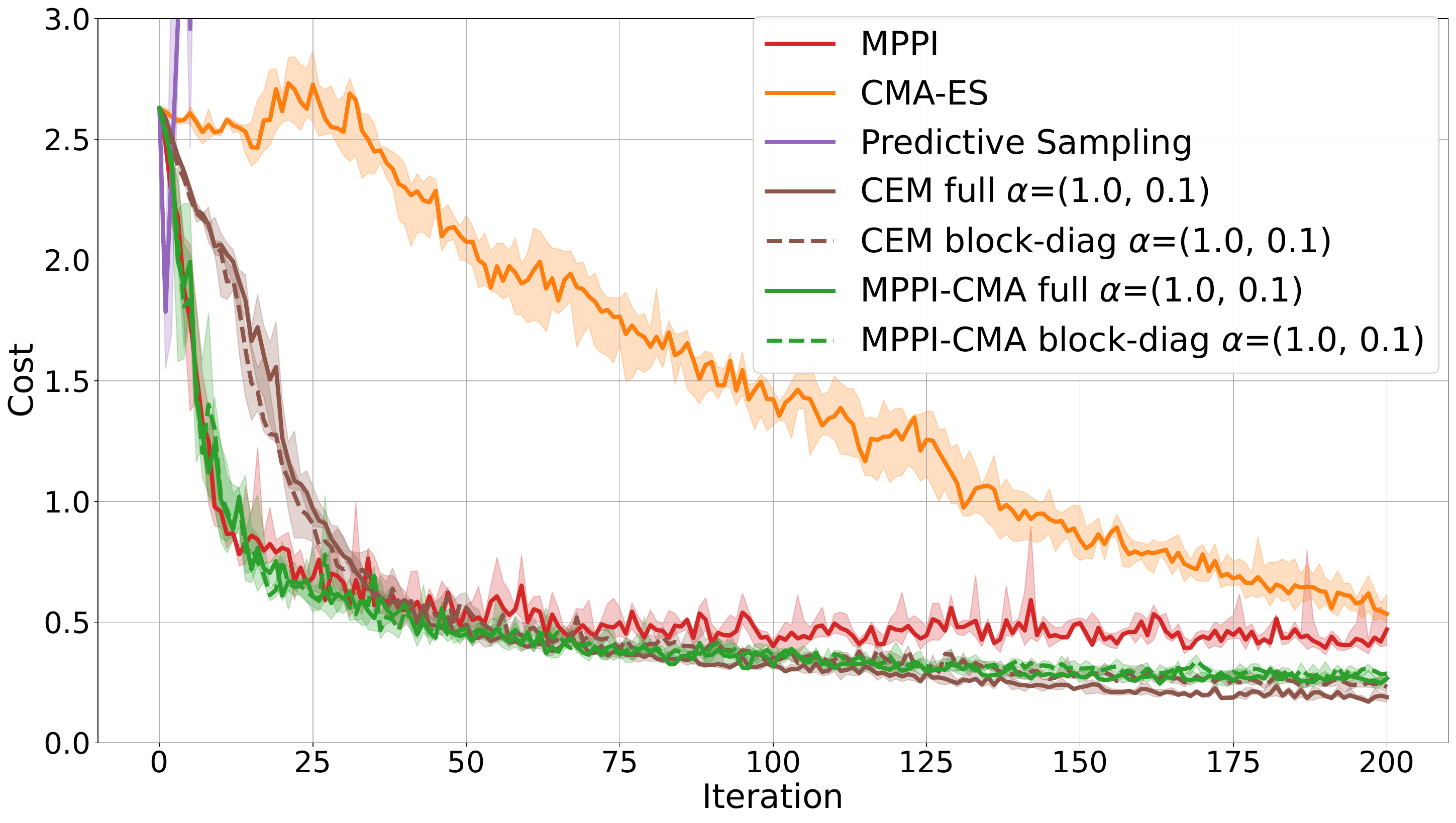}
  \caption{Go2Walk}\end{subfigure}
\caption{Median cost according to the number of iterations for different test problems. The median is taken over 6 seeds. MPPI-CMA and CEM perform the best: CEM sometimes reaches lower minima while MPPI-CMA sometimes decreases the cost faster during the first few iterations.}
\label{fig:TO_benchmark}
\end{figure*}

Figure~\ref{fig:TO_benchmark} provides additional results for the TO benchmark. Specifically, we show how block-diagonal CMA compares to the default CMA (which relies on the full matrix). In addition, we illustrate the advantage of having different step-sizes for the mean and the covariance.


\begin{figure*}
\centering
\begin{subfigure}{0.49\linewidth}\centering
  \includegraphics[width=\linewidth]{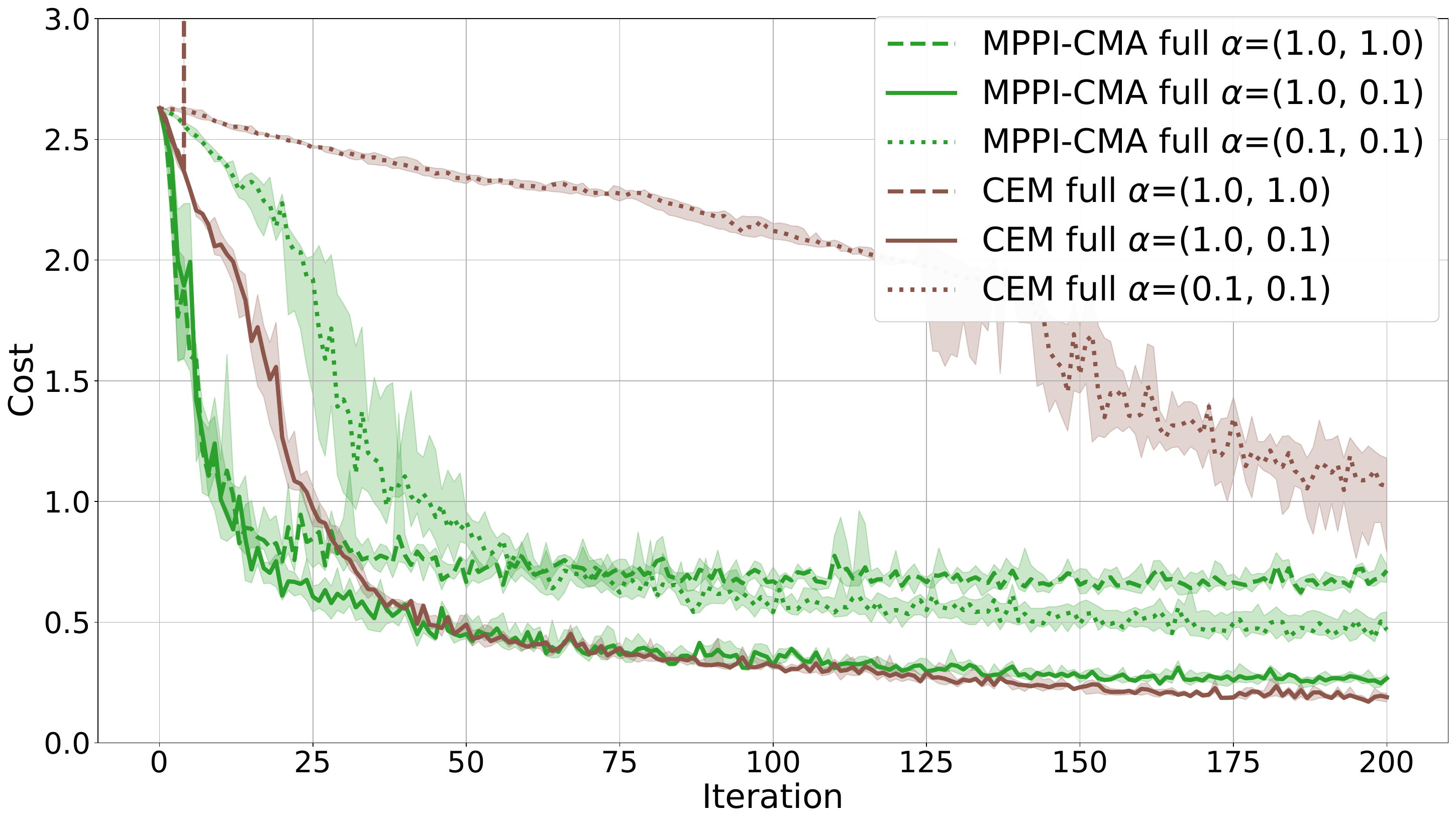}
  \caption{Go2Walk}\end{subfigure}\hfill
\begin{subfigure}{0.49\linewidth}\centering
  \includegraphics[width=\linewidth]{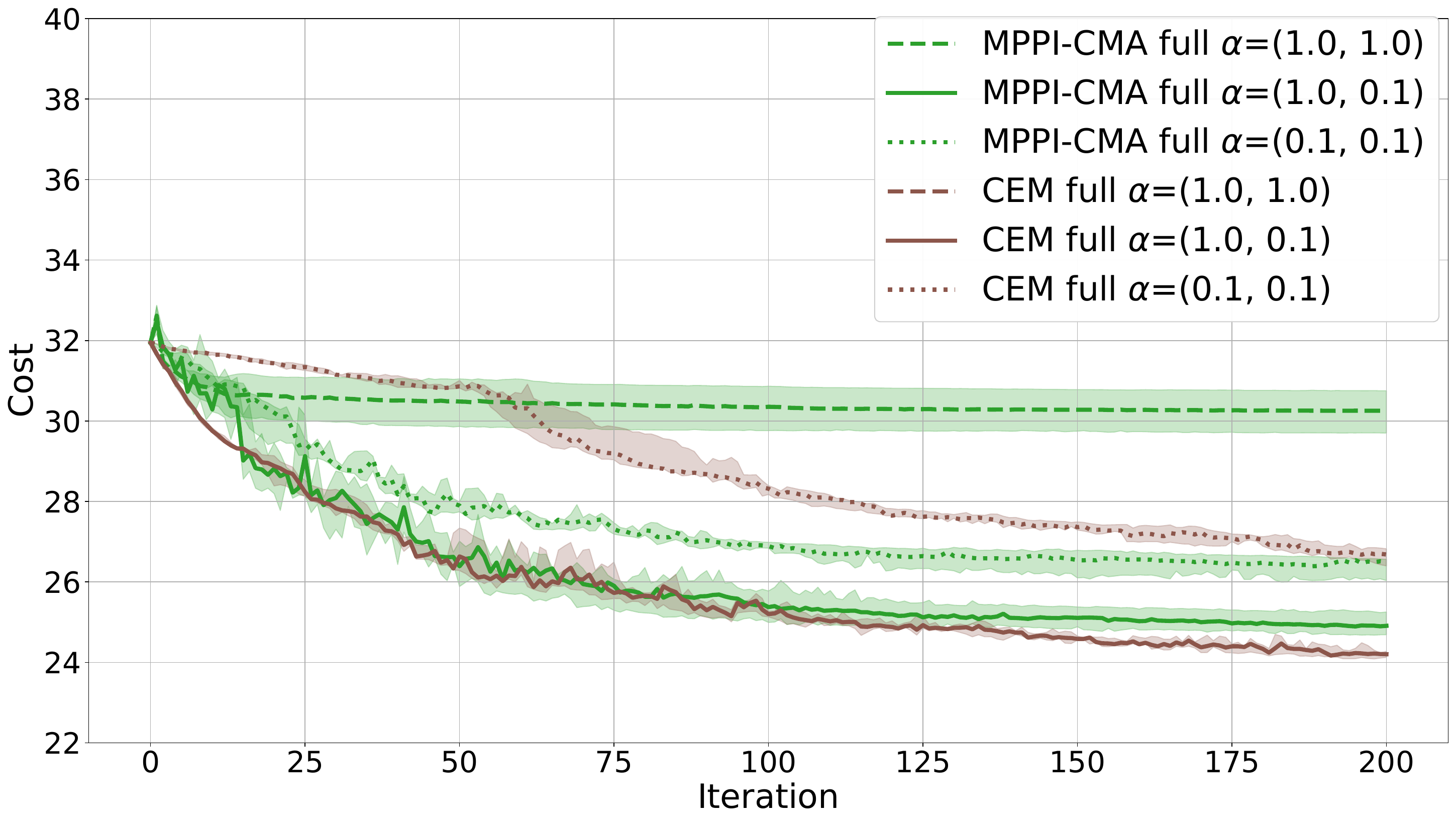}
  \caption{HumanoidStandup}\end{subfigure}
\caption{Median cost according to the number of iterations for MPPI-CMA and CEM with full covariance update. The median is taken over 6 seeds. Covariance adaptation needs two distinct learning rates: for both MPPI-CMA and CEM, $(\alpha_{m}, \alpha_{\Sigma} )=(1,0.1)$ outperforms a single shared rate $(1,1)$ or $(0.1,0.1)$ on both Go2Walk and HumanoidStandup. }
\label{fig:distinct_lr}
\end{figure*}

\section{Additional RL results}~\label{appendix:RLresults}

\begin{figure*}
\centering
    \includegraphics[width=0.75\linewidth]{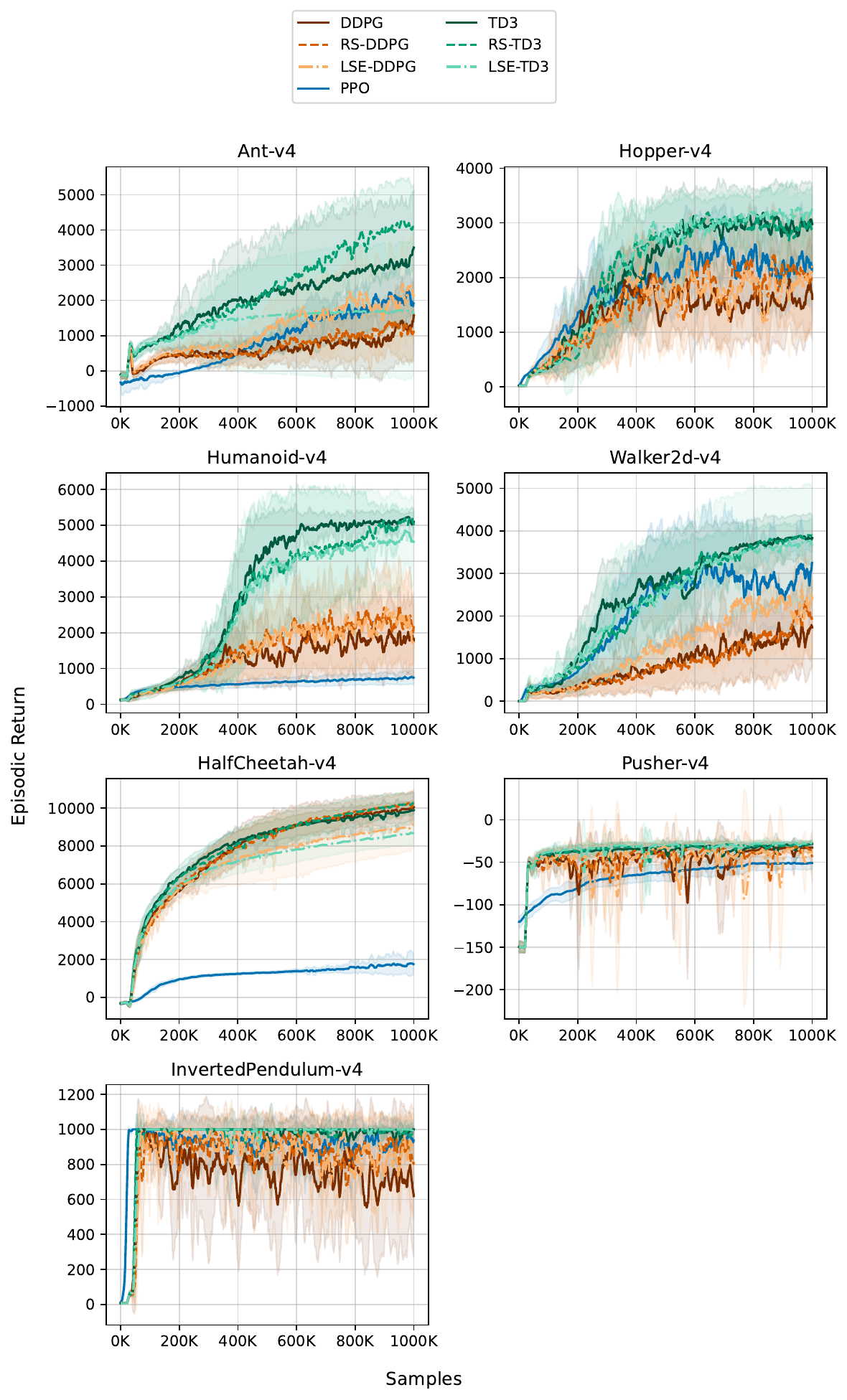}
    \caption{Episodic return for each environment.}
    \label{fig:rl-compare}
\end{figure*}

Figure~\ref{fig:rl-compare} shows the episodic return for each test environment.

\end{document}